\PassOptionsToPackage{table}{xcolor}
\documentclass[a4paper,fleqn]{cas-dc}

\usepackage[numbers]{natbib}
\usepackage{xcolor}
\usepackage{xurl}
\usepackage{booktabs}
\usepackage{tabularx}
\usepackage{hyperref}
\definecolor{tablegray}{gray}{0.9}
\usepackage{cleveref}
\usepackage{array}
\usepackage{ragged2e}
\usepackage{hyphenat}
\usepackage{enumitem}

\begin{document}
\let\WriteBookmarks\relax
\def\floatpagepagefraction{1}
\def\textpagefraction{.001}

\shorttitle{Agent Skill Harnessing}    

\shortauthors{Xia et al.}  

\title [mode = title]{Harnessing Agent Skills: Architectural Patterns and a Reference Architecture for Skill-Mediated LLM Agents}  



\author[1,2,3]{Boming Xia}[
    orcid=0009-0003-7385-4023
]
\ead{boming.xia@adelaide.edu.au}
\credit{Conceptualization, Methodology, Writing - Original draft, Writing - Review \& Editing}

\author[3,4]{Liming Zhu}[
    orcid=0000-0001-5839-3765
]
\ead{liming.zhu@csiro.au}
\credit{Supervision, Conceptualization, Writing - Review \& Editing}

\author[3]{Zhenchang Xing}[
    orcid=0000-0001-7663-1421
]
\ead{zhenchang.xing@csiro.au}
\credit{Supervision, Conceptualization, Writing - Review \& Editing}

\author[3,4,1]{Qinghua Lu}[
    orcid=0000-0002-9466-1672
]
\ead{qinghua.lu@csiro.au}
\credit{Supervision, Conceptualization, Writing - Review \& Editing}

\author[1,2]{Dino Sejdinovic}[
    orcid=0000-0001-5547-9213
]
\ead{dino.sejdinovic@adelaide.edu.au}
\credit{Supervision, Writing - Review \& Editing}

\author[3,4]{Xiwei Xu}[
    orcid=0000-0002-2273-1862
]
\ead{xiwei.xu@csiro.au}
\credit{Writing - Review \& Editing}

\affiliation[1]{organization={Responsible AI Research (RAIR) Centre},
    city={Adelaide},
    state={SA},
    country={Australia}}

\affiliation[2]{organization={Adelaide University},
    city={Adelaide},
    state={SA},
    country={Australia}}

\affiliation[3]{organization={CSIRO},
    city={Sydney},
    state={NSW},
    country={Australia}}

\affiliation[4]{organization={University of New South Wales},
    city={Sydney},
    state={NSW},
    country={Australia}}



\begin{abstract}
CONTEXT: Agent skills package reusable behavioural guidance and resources that large language model agents load or apply during a run. Within the agent harness, agent skill harnessing comprises connected architectural responsibilities that govern which skills enter a run, which actions they can cause, and how their use is recorded and informs revision.
OBJECTIVES: We identify architectural patterns for agent skill harnessing and organise their responsibilities into a reference architecture that makes these connections explicit.
METHODS: A multivocal literature review covered 36 practitioner systems and 51 academic papers; we consolidated practices into patterns and derived the reference architecture from patterns, source evidence, and architectural reasoning. We evaluated the reference architecture by mapping eight systems and applying it in Deep Agents and Hermes Agent.
RESULTS: Ten patterns cover skill supply and activation, skill declarations and execution authority, verification and recording of skill use, and run evidence feeding repair and governed revision. The reference architecture organises their responsibilities into four layers (Supply Chain, Mediation, Execution Control, Evidence \& Feedback) and two cross-cutting substrates, separating permission determination from enforcement and routing persistent revisions through acceptance. Mapping eight systems showed the first three layers broadly realised and 14 of 15 partial or unestablished ratings in Evidence \& Feedback. Applying the architecture in Deep Agents and Hermes Agent turned requirements into checks that exposed a shell route outside the file-tool rules and a preview--apply mismatch.
CONCLUSION: The patterns and reference architecture support comparing implementations and turning requirements into responsibility allocations and checks.
\end{abstract}




 


\maketitle
\hypersetup{pdfauthor={Boming Xia, Liming Zhu, Zhenchang Xing, Qinghua Lu, Dino Sejdinovic, Xiwei Xu}}

\section{Introduction}
\label{sec:introduction}

\begin{sloppypar}
Agentic AI systems allocate behavioural capability across prompts, retrieval, memory, workflows, callable tools, and externalised behavioural artefacts, alongside the capabilities encoded in model weights~\citep{anthropic2024buildingeffectiveagents,zhou2026externalization}. \emph{Agent skills} are one such artefact form: reusable units of behavioural guidance or intervention logic stored outside model weights, together with associated metadata and resources, for use by a large language model (LLM) agent. They include instruction-based packages and programmatic forms that intervene in the agent loop~\citep{anthropicEqquipAgentSkills,agentskillsOverview,liu2026harnessing}. Externalising these resources allows them to be inspected and revised independently of the model. It also places decisions about their availability, selection, use, and evolution in the surrounding software system. Because these decisions span components and affect runtime control, trust boundaries, and traceability, their interaction is an architectural concern.
\end{sloppypar}

\begin{figure*}
\centering
\includegraphics[width=0.85\linewidth]{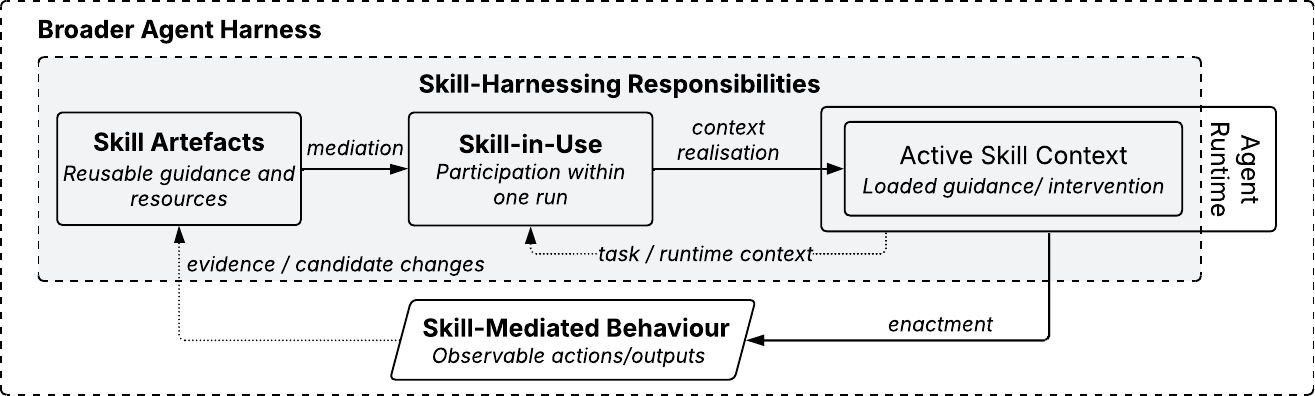}
\caption{From skill artefact to skill-mediated behaviour. Mediation decides whether an artefact participates in a run as skill-in-use; context realisation supplies its guidance, resources, or intervention as active skill context in the agent runtime; enactment produces the resulting behaviour, whose evidence can yield candidate changes to the artefact.}
\label{fig:artefact2Behaviour}
\end{figure*}

The central issue is how a reusable skill artefact participates in a particular agent run. Participation depends on the task, available context, mechanisms of the host (the agent application or platform), and applicable policies. We call this run-specific participation \emph{skill-in-use}, and the associated observable actions and outputs \emph{skill-mediated behaviour} (\Cref{fig:artefact2Behaviour}).
Skill participation also interacts with the permissions governing execution. The Agent Skills specification includes experimental \texttt{allowed-tools} pre-approval metadata, with host-dependent support, and GitHub Copilot CLI documents pre-approval of listed tools~\citep{agentskillsSpec,githubCopilotCliSkills}. What an agent may do under a skill therefore depends on the applicable permissions as well as on the skill’s instructions.

These participation and authorisation decisions connect to admission, execution control, evidence capture, and persistent change. We use \emph{agent skill harnessing} for the architectural responsibilities and mechanisms that govern skill supply, runtime participation, and executable effects, and link them to evidence and revision. It sits within the broader harness that coordinates context, tools, execution, and observation~\citep{zhou2026externalization}. \Cref{sec:conceptual-foundation} develops the scope and definitions.

General architectures for LLM agents already cover guardrails, runtime recording, and provenance~\citep{lu2023responsiblearchitecture} and registries, adapters, and evaluators~\citep{liu2024agentdesignpatterns}. Skill-focused work covers selection, permission-checked binding, and lifecycle governance in reviews and a lifecycle-staged reference architecture~\citep{zhou2026externalization,xu2026agentskills,badhe2026agenticskills}.
These works organise mechanisms by topic or lifecycle stage. We synthesise documented practices into architectural patterns and a reference architecture (RA) organised by the objects that skill decisions govern. The RA specifies how a skill's declarations enter the host's permission decision, how that decision is enforced on each route that runs the skill's code, and how runs and revisions are traced back to the skill, supporting comparison of implementations and concrete requirement checks.

We conducted a multivocal literature review~\citep{garousi2019guidelines} covering 36 practitioner-documented systems and 51 academic papers, with searches in May 2026. Extracted practices informed ten patterns: five core patterns for the central concerns (participation, authority, verification, provenance, and feedback) and five supporting patterns for narrower decisions. We constructed the pattern-oriented RA by adapting Galster and Avgeriou's method for empirically grounded RAs~\citep{galster2011empirically}, combining pattern solutions, underlying practices, and architectural reasoning. The RA has four responsibility layers (Supply Chain, Mediation, Execution Control, and Evidence \& Feedback) and two cross-cutting substrates (Policy/Configuration and Identifier).

We evaluated the RA through mappings of eight systems and focused technical applications in \href{https://github.com/langchain-ai/deepagents}{Deep Agents} (LangChain) and \href{https://github.com/NousResearch/hermes-agent}{Hermes Agent} (Nous Research), hereafter Hermes. The mappings compare how documented systems realise each responsibility and under which conditions; the applications examine whether selected responsibilities and dependencies translate into concrete configurations whose consequences can be checked. The paper makes two main contributions:
\begin{itemize}
    \item \textbf{Pattern catalogue.} Ten architectural patterns describing skill-centred problems, solution structures, forces, and trade-offs, with representative source support.
    \item \textbf{Reference architecture and its evaluation.} A logical organisation of skill-harnessing responsibilities, their interactions, and variation points, distinguishing skill-specific decisions from shared harness mechanisms. Mappings of eight systems locate most partial or unestablished realisations in Evidence \& Feedback; two applications turn RA dependencies into checks that exposed gaps in documented host controls.
\end{itemize}

\section{Conceptual Foundation}
\label{sec:conceptual-foundation}

The patterns and RA rest on three objects: a persistent skill artefact, its participation within a run, and the behaviour observed during that participation (\Cref{fig:artefact2Behaviour}).

\begin{table*}
\caption{Related forms and their architectural roles.}
\label{tab:adjacent-forms}
\centering
\footnotesize
\begin{tabularx}{\linewidth}{@{}>{\raggedright\arraybackslash}p{0.20\linewidth}>{\raggedright\arraybackslash}X>{\raggedright\arraybackslash}X@{}}
\toprule
\textbf{Form} & \textbf{Contribution and runtime role} & \textbf{Relation/Distinction to skill artefacts} \\
\midrule
\rowcolor{tablegray}
Model weights
& Parametric capabilities used during inference
& Internal to the model; skill artefacts are represented outside them \\
\addlinespace
Reusable prompts, instruction files
& Textual guidance loaded into model context
& Can implement a skill artefact when a host supplies it as an identifiable, reusable behavioural unit \\
\addlinespace
\rowcolor{tablegray}
Retrieved context and memory
& Facts, experience, or behavioural guidance supplied as context
& Can supply skill artefacts as well as ordinary contextual information \\
\addlinespace
Callable tools, code, scripts
& Callable operations executed by a runtime or service
& Can implement a skill's intervention or helpers \\
\addlinespace
\rowcolor{tablegray}
Workflows, plans
& Task structure and coordinated steps used by an agent or orchestrator
& Can organise skill use within a larger execution process \\
\addlinespace
Plugin frameworks
& Extension and distribution mechanisms
& Can distribute skills alongside other extensions \\
\addlinespace
\rowcolor{tablegray}
Model Context Protocol (MCP)~\citep{mcp2026spec}
& Interfaces to operations and resources
& Can expose capabilities used by skills; its Skills extension can also supply skill artefacts~\citep{mcpSkillsExtension} \\
\addlinespace
\textbf{Skill artefacts}
& Reusable guidance or intervention logic loaded or applied by the host, with associated resources
& One host-supplied unit can combine guidance, executable resources, and, in some formats, permission declarations \\
\bottomrule
\end{tabularx}
\end{table*}

\subsection{Skill Artefact, Skill-in-Use, and Skill-Mediated Behaviour}
\label{subsec:externalisation}

\textbf{Skill artefact.} A skill artefact is an identifiable, persistent unit of behavioural guidance or intervention logic stored outside model weights, together with any associated metadata and resources. The host supplies or applies it as an individual, reusable unit through selection, invocation, an activation condition, or configuration. Its representation can be textual~\citep{agentskillsSpec}, executable~\citep{liu2026harnessing}, or both.

\textbf{Skill-in-use.} Skill-in-use is the participation of a skill artefact in a particular run, such as a task or session whose boundaries the analysis specifies. Participation can begin through invocation or preloading, or carry over from an earlier run. Within the agent runtime, it is represented by the \emph{active skill context}: the loaded instructions, available resources, or intervention state.

\textbf{Skill-mediated behaviour.} Skill-mediated behaviour comprises observable actions and outputs linked to skill-in-use by run evidence, such as tool calls, generated files, or changes to proposed actions.

Anthropic's PDF skill illustrates the three objects~\citep{anthropicEqquipAgentSkills,anthropicPdfSkill}. Its package, which includes \texttt{SKILL.md}, \texttt{forms.md}, and helper scripts, is the skill artefact. In the published walk-through, Claude reads \texttt{SKILL.md} through a Bash tool call and then reads \texttt{forms.md}. This staged loading is the skill-in-use for the task, and the loaded instructions form its active skill context. The reads themselves are skill-mediated behaviour: they show that the files were loaded, not that Claude carried out their PDF-processing instructions.

\subsection{Skill Artefacts and Related Forms}
\label{subsec:adjacent-forms}

Skill artefacts interact with several related forms, and one implementation can combine their roles (\Cref{tab:adjacent-forms}). Hosts supply skill artefacts in different ways: Claude Code can load named skills into a configured subagent at startup~\citep{anthropic2026claudesubagents}, and OpenHands supplies path-triggered rules and repository instructions such as \texttt{AGENTS.md} through its skill mechanisms~\citep{openhands2026skillsContext}.
Managed instruction files are therefore skill artefacts whether they are loaded on demand or, like \texttt{AGENTS.md}, in every run.

\subsection{Architectural Concerns in the Artefact-to-Use Transition}
\label{subsec:concerns}

Each step from stored artefact to observed change raises its own concerns. Supply raises trust and admission, which the supporting patterns address (\Cref{sec:supporting-patterns}). The five concerns below arise once a skill enters a run; each refines a concern shared with broader harnesses.

\textbf{Runtime participation.} A skill's descriptor, such as its name and description, can be exposed before activation loads its instructions or applies its intervention, and supporting resources may be accessed later still~\citep{liu2026harnessing,agentskillsSpec}. These stages distinguish availability and selection from activation and use.

\textbf{Skill declarations and execution authority.} Execution authority means the permissions applicable to an action, as enforced by host and resource controls. Naming a capability in guidance does not grant permission to use it, but hosts can interpret skill declarations as permission inputs.
In Claude Code's documented \texttt{commit} example, the \texttt{allowed-tools} field lists \texttt{git add}, \texttt{git commit}, and \texttt{git status}. When host policy honours it, these commands are pre-approved only for the invoking turn, while the skill's instructions can remain in context~\citep{anthropic2026claudeskills}. Instruction and permission lifetimes can therefore differ.

\textbf{Enactment and verification.} Verification criteria must match the enactment mechanism and the claimed outcome: loading guidance does not show that the agent followed it, and executing an intervention does not show that the task succeeded.

\textbf{Run-specific behavioural provenance.} A skill inventory shows which skills were available, whereas provenance records which of them participated in a particular run. It links artefact identity and version with participation, permission decisions, and observed behaviour. Such links establish association, not causal contribution to task outcomes.

\textbf{Run-to-artefact feedback.} Run-local repair changes skill-in-use within the current run, whereas persistent evolution revises the artefact supplied to future runs. Because a persistent revision can also reach other users of the artefact, its review must consider consequences beyond the run that prompted it.

The patterns in \Cref{sec:patterns} address these concerns, and the RA in \Cref{sec:reference-architecture} organises their responsibilities and their connections to the surrounding harness.

\section{Methodology}
\label{sec:methodology}
\begin{figure*}
    \centering
    \includegraphics[width=0.9\linewidth]{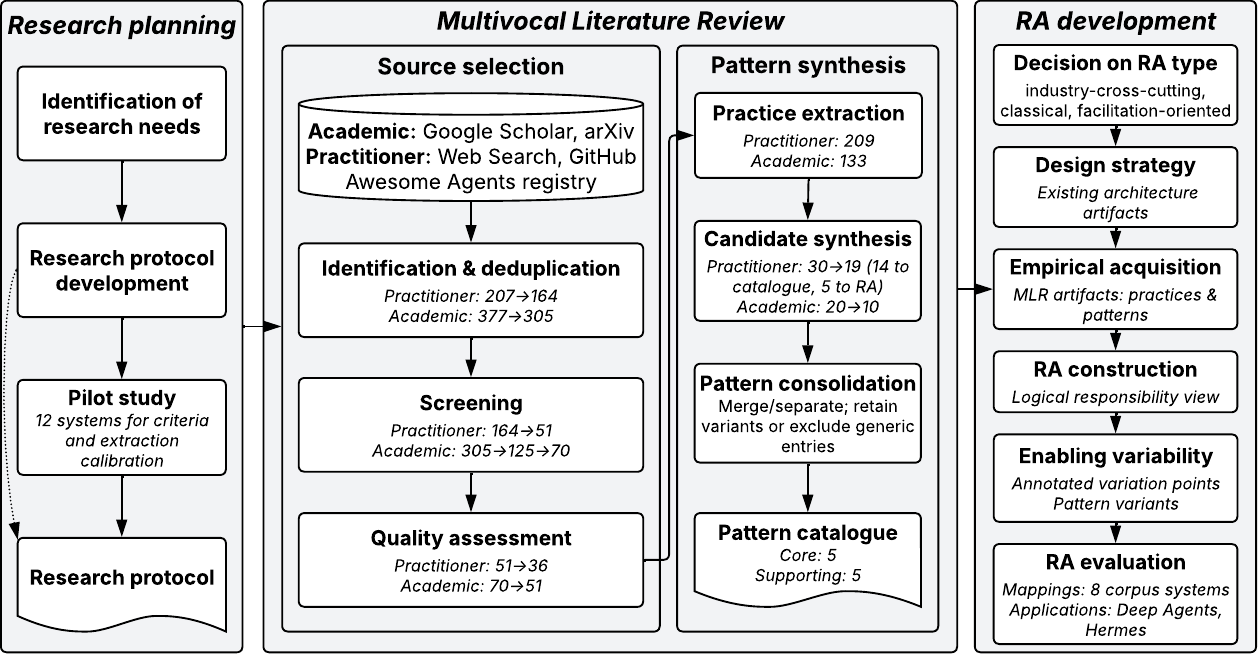}
    \caption{Methodology, with RA development adapted from the six-step method of Galster and Avgeriou~\citep{galster2011empirically}. Arrows give counts before and after each stage.}
    \label{fig:methodology}
\end{figure*}

We developed the pattern catalogue (\Cref{tab:catalogue}) and RA through three stages (\Cref{fig:methodology}). First, we planned the review and piloted it: two researchers independently assessed a pilot sample of 12 systems to refine the assessment criteria and extraction schema, resolving disagreements through discussion, and preliminary candidate patterns from the pilot informed later coding. Second, a multivocal literature review (MLR), guided by established methods~\citep{garousi2019guidelines,kitchenham2007guidelines}, identified documented skill-harnessing practices and consolidated them into patterns. Third, we constructed an empirically grounded RA from patterns, source evidence, and architectural reasoning (\Cref{sec:methodology-ra}). After the pilot, the first author assessed the remaining sources, constructed the candidate patterns, consolidated the catalogue, and developed the RA. Ambiguous cases were discussed among coauthors and resolved by consensus.

\subsection{Multivocal Literature Review}
\label{sec:methodology-mlr}

The MLR combined first-party technical documentation of agent systems with academic accounts of skill-related practices. The supplementary materials provide the source, practice, synthesis, and RA derivation records, and the application protocols, drivers, and results\footnote{\url{https://github.com/bomingxia/harnessing-agent-skills-supplement}}.

\subsubsection{Source Identification}

For practitioner sources, we screened 137 entries from the \href{https://github.com/kyrolabs/awesome-agents}{\textit{Awesome Agents}} registry and 13 results of the web query \texttt{(Agent OR Agentic) AND (Skill OR Skills)}. Follow-up searches targeting named systems and skill-documentation terms identified 54 further records; following first-party links added 3. Searches were conducted on 13 May 2026. Search results identified systems; extraction used first-party documentation and repositories. Deduplication and removal of non-system records reduced the 207 records to 164.

For academic sources, we searched Google Scholar and arXiv on 8 May 2026 for records from 2025 onward. The Google Scholar query was \texttt{allintitle:(LLM OR LLMs OR "Large Language Model" OR "Large Language Models" OR Agent OR Agents OR Agentic OR Harness) AND (Skill OR Skills)}, with an equivalent title-field query in arXiv. This precision-oriented strategy targeted contemporary agent skills: even the title-restricted searches returned uses of ``skill'' for human training, robotics, and model competence. Google Scholar returned 165 records and arXiv 207; targeted snowballing from four seed papers added 5. Deduplication retained 305 unique academic records.

\subsubsection{Eligibility and Detailed Assessment}

Both practitioner and academic streams applied the same eligibility criteria. A source was eligible if it described a reusable, externally represented, agent-facing skill artefact, as characterised in \Cref{sec:conceptual-foundation}, together with a mechanism that managed, governed, recorded, or revised its participation in a run. Sources describing only generic orchestration, tools, plugins, package management, or prompt templates without such a managed artefact were excluded, as were skill literatures concerned with agent-internal learned policies or action repertoires.
Eligible sources needed sufficient technical detail to identify a concrete mechanism and its architectural responsibility. Practitioner evidence had to be first-party.

For practitioner sources, initial keyword screening was followed by skill-signal triage: checking first-party README or documentation pages for skill-related terminology tied to a reusable agent-facing artefact. This retained 51 practitioner sources for detailed assessment: 22 from the registry and 29 from web searches. Detailed assessment considered fit with agent-system skill harnessing, duplicate product lineages, maintenance, adoption, and architectural detail beyond already retained sources. Repository activity within the preceding six months informed maintenance assessment; current first-party technical documentation could also establish currency. We retained 36 practitioner systems.

For academic sources, title and abstract screening retained 125 of the 305 records for eligibility checking: 70 advanced to full-text assessment, 44 became background records without practice extraction, and 11 were excluded. Full-text assessment rated technical specificity, practice extractability, and evidence quality (implementation and evaluation detail), and considered study type and transferability beyond one application domain; it retained 51 papers for extraction, classified 17 as background, and excluded 2.

\subsubsection{Practice Extraction}

A practice is a source-supported mechanism or architectural decision concerning a skill artefact or its use. Each practice record linked a located source passage to a practice description, a descriptive mechanism category, and boundary or audit notes. The practitioner and academic streams yielded 209 and 133 practice records.

\subsubsection{Pattern Synthesis and Consolidation}
\label{sec:methodology-synthesis}

Within each stream, we grouped practices by architectural problem and responsibility-level solution, producing 30 practitioner-stream and 20 academic-stream candidate patterns from the 342 practices. Whether to retain, merge, or exclude each candidate was a qualitative judgement on recurrence, reusability, forces and consequences, traceable support, and what the skill artefact changed about the decision, with the reasons recorded.

The practitioner stream kept 19 of its 30 candidates: 14 for the catalogue and five that inform RA responsibilities, supplying the basic artefact unit and enabling functions such as supply and enforcement (\Cref{tab:ra-responsibilities}). The other 11 were not retained: five adjacent or general mechanisms, of which generic permission and sandbox policies later inform Runtime Action Control; five non-pattern proposals, such as command syntax; and one inventory-and-preview component. The academic stream consolidated its 20 candidates into ten intermediate entries. The 14 practitioner candidates and the ten academic entries were then consolidated into the catalogue. Within and across streams, we merged equivalent problem--solution structures, such as the activation candidates; separated distinct responsibilities, such as run-local repair and persistent evolution; and absorbed narrower features as variants, such as parameterised invocation.

The catalogue contains ten patterns: five core and five supporting. Representative support and qualifications are reported with each pattern in \Cref{sec:patterns}, and \Cref{sec:ra-overview} traces one candidate from source practices to the RA.

\subsection{Reference Architecture Construction}
\label{sec:methodology-ra}

We adapted the six-step method of Galster and Avgeriou~\citep{galster2011empirically} by placing the pattern catalogue between the documentary evidence and the logical architecture description. The method distinguishes the empirical foundation established through Steps~1--5 from the evaluation in Step~6.

\textbf{Step 1 (Decision on RA type).}
In usage context, the RA is industry-cross-cutting: it organises common skill-harnessing responsibilities for agent-system designers and implementers across application sectors, organisations, and platforms. It is classical because it draws on experience from existing systems, and facilitation-oriented because it provides guidance for architectural design and analysis~\citep{angelov2009classification}.

\textbf{Step 2 (Selection of design strategy).}
The design strategy is generalisation from existing architectural evidence. Patterns supply reusable problem--solution structures; the underlying practices ground these abstractions and inform the enabling functions needed for their interaction. This pattern-oriented strategy is consistent with earlier reference architecture work~\citep{lu2023responsiblearchitecture}.

\textbf{Step 3 (Empirical data acquisition).}
RA construction used the extracted practices, their pattern solutions, and documented shared harness mechanisms (\Cref{sec:methodology-synthesis,sec:supporting-patterns}), with the concerns in \Cref{subsec:concerns} as architectural requirements.

\textbf{Step 4 (RA construction).}
We derived responsibilities from pattern solutions and practice records and grouped them into blocks around a shared object or decision boundary: scope resolution and eligibility share a block because both determine the selectable pool, whereas permission determination and enforcement, like evidence capture and verification, form separate blocks. Functions that pattern solutions presuppose, such as artefact supply, form enabling blocks; rules and identifiers used across blocks form substrates. Layers group blocks by the object they govern: supplied artefacts, runtime participation, action authority, and run evidence and change. Interactions link blocks through documented or inferred input--output dependencies. Block names state the grouped decisions, as in Resolution \& Admission, or the enabling function, as in Distribution. \Cref{sec:reference-architecture} specifies these conventions as a \emph{Responsibility Viewpoint} following ISO/\allowbreak{}IEC/\allowbreak{}IEEE 42010~\citep{isoiecieee42010} and presents the resulting Responsibility View. Derivation records, compiled after construction, link each allocation and dependency to its source evidence and reasoning.

\textbf{Step 5 (Enabling variability).}
We derived variation points from documented pattern variants and differences among systems. These are represented as textual annotations describing alternative realisations of responsibilities, while preserving their dependencies (\Cref{sec:useVariationLimits}).

\textbf{Step 6 (RA evaluation).}
We assessed the RA through documentary mappings of eight review-corpus systems and focused technical applications, reported in \Cref{sec:evaluation}.

\section{Architectural Patterns for Skill Harnessing}
\label{sec:patterns}

\Cref{tab:catalogue} summarises the ten patterns. The five core patterns address the five concerns of \Cref{subsec:concerns} (\Cref{sec:pattern_progressiveskillactivation,sec:pattern_skillauthority,sec:pattern_verifiableskillcontract,sec:pattern_skillbom,sec:pattern_skillagentco}); the five supporting patterns address narrower decisions, including trust and admission, that realise or complement them. Each core pattern states its intent, problem, forces, solution, variants, consequences, and evidence.
The pattern descriptions also draw on current host documentation and sources outside the review corpus to illustrate mechanisms and variants.

\begin{table*}
\centering
\caption{Catalogue of skill-harnessing architectural patterns.}
\label{tab:catalogue}
\small
\begin{tabularx}{\textwidth}{@{}>{\RaggedRight\arraybackslash}p{0.24\textwidth}>{\RaggedRight\arraybackslash}X@{}}
\toprule
\textbf{Pattern} & \textbf{Problem and solution in brief} \\
\midrule
\multicolumn{2}{c}{\textbf{Core patterns}} \\
\addlinespace[2pt]
\rowcolor{tablegray}
Progressive Skill Activation
  & Available skill artefacts may influence behaviour or consume context if their guidance is exposed too early in a run. Hosts expose compact descriptors first and load full guidance and resources only on activation. \\
Skill--Execution Authority Separation
  & Skill declarations can affect permissions when interpreted by a host. Host and resource policy decides each action's permissions; skill declarations are inputs to that decision, separate from skill participation. \\
\rowcolor{tablegray}
Verifiable Skill Contract$^\dagger$
  & Loading or invoking a skill does not show that it was followed or succeeded. A verifier checks skill-scoped criteria against run evidence and records a verification status. \\
Runtime Skill Bill of Materials$^\dagger$
  & Repository inventories do not identify which skill artefacts and versions participated in a run. A skill-centred record links participation decisions to run evidence. \\
\rowcolor{tablegray}
Skill--Agent Co-Evolution Loop
  & Run evidence may suggest skill changes. Candidate changes require validation before entering the shared skill supply. \\
\midrule
\multicolumn{2}{c}{\textbf{Supporting patterns}} \\
\addlinespace[2pt]
\rowcolor{tablegray}
Skill Eligibility Gate
  & Supplied skills require assessment for admission to the selectable pool. A gate applies validation, trust, or compatibility conditions. \\
Skill Scope Cascade
  & Skills defined at several scopes, such as organisation, user, and project, can expose unintended guidance or resources. Visibility and precedence rules determine which skills are visible and resolve competing artefacts. \\
\rowcolor{tablegray}
Skill Resource Materialisation Boundary
  & Reading a skill's files and running its helpers have different costs and permissions. An explicit materialisation decision identifies which resources are supplied and which helper operations occur. \\
Skill Extension Bundle
  & Guidance and its supporting material can be distributed separately. A deployable bundle gives them coordinated distribution, compatibility, and review boundaries. \\
\rowcolor{tablegray}
Runtime Skill Repair$^\dagger$
  & A skill may fail verification or skill-scoped enactment criteria mid-run. Bounded adaptation of skill-in-use lets the run continue without changing the shared artefact. \\
\bottomrule
\end{tabularx}

\par\smallskip
\begin{minipage}{\linewidth}
\footnotesize
\raggedright
$^\dagger$Known uses are mainly research systems; practitioner systems document partial mechanisms.
\end{minipage}
\end{table*}

\subsection{Progressive Skill Activation}
\label{sec:pattern_progressiveskillactivation}

\begin{figure}
    \centering
    \includegraphics[width=\linewidth]{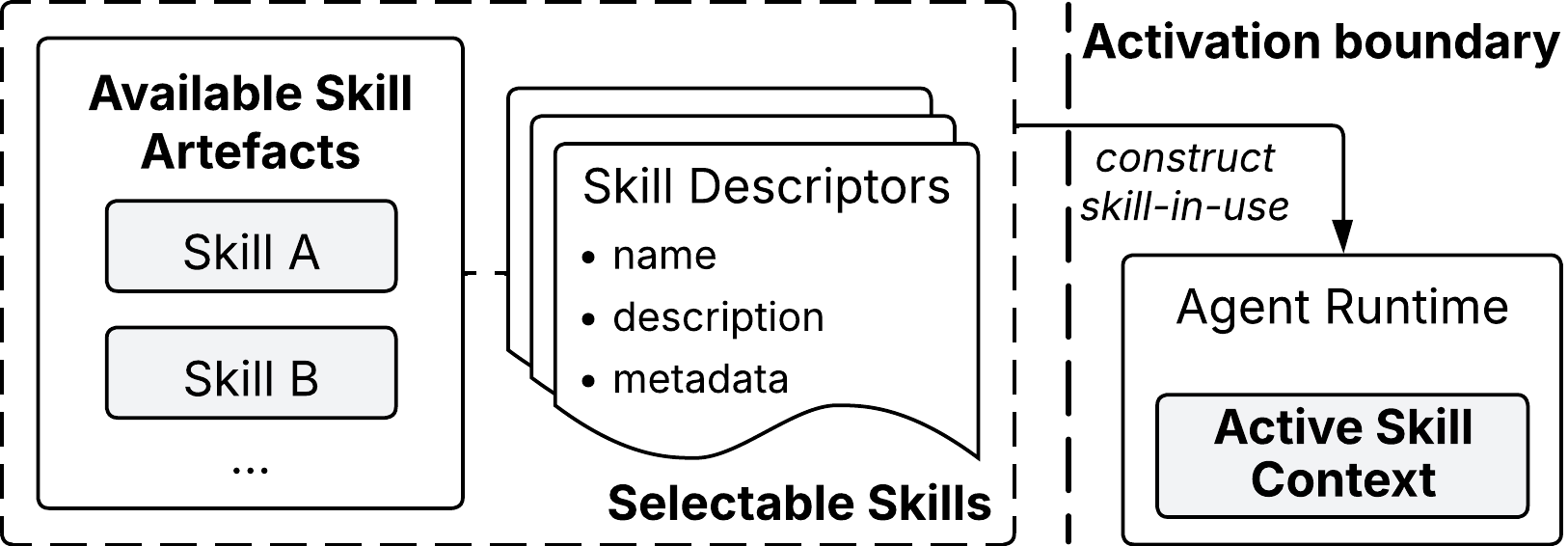}
    \caption{Progressive Skill Activation: in the descriptor-first realisation shown, compact descriptors support selection and activation supplies instructions as active skill context.}
    \label{fig:progressive-skill-activation}
\end{figure}

\noindent\textbf{Intent.}
Stage the transition from available skill artefacts to active skill-in-use, exposing or applying what is required at each stage (\Cref{fig:progressive-skill-activation}).

\noindent\textbf{Context and problem.}
Agent systems may expose many skills from various sources. Loading all skill bodies into a run can consume context unnecessarily and introduce irrelevant guidance. Hiding all skill information until invocation creates the opposite problem: the agent or user cannot identify relevant reusable behaviour. The architectural problem is to provide enough information for selection while controlling when full guidance, resources, or interventions participate.

\noindent\textbf{Forces.}
\emph{(i) Discoverability vs.\ premature full exposure.}
A skill must be visible enough to support selection, but exposing full guidance too early consumes context~\citep{perplexity2026designingrefiningskills} and may influence the run before the skill is needed.
\emph{(ii) Coverage vs.\ activation precision.}
Large skill inventories improve reuse coverage but also increase selection ambiguity and the risk of activating irrelevant or conflicting guidance.
\emph{(iii) Traceability vs.\ operational simplicity.}
Inspecting skill influence supports review, repair, and revocation, but finer-grained records increase maintenance cost.

\noindent\textbf{Solution.}
Activate skills through staged architectural states. A skill artefact is \emph{available} when present in a source the system can inspect. It is \emph{selectable} when the host exposes it for selection, commonly through a compact descriptor. It is \emph{active} when invocation, configuration, or an activation condition supplies its guidance or applies its intervention in the run.
Where supporting material is required, activation remains separate from resource materialisation.
Descriptor-first disclosure keeps candidate skills discoverable while delaying full instruction loading. Configured preloading can activate a named skill at startup, and programmatic skills can activate through a predicate.

\noindent\textbf{Variants.}
Variants differ in who initiates activation.
In \emph{explicit activation}, a user, command, prompt mention, or interface action requests that a selectable skill become active.
In \emph{policy-mediated activation}, the system activates selectable skills according to configuration, scope, triggers, or applicability predicates.
In \emph{agent-proposed activation}, the agent proposes a skill from exposed descriptors, while the system decides whether activation is permitted.
Activation may occur at run start or be re-applied as a task proceeds.

\noindent\textbf{Consequences.}
\emph{(i) Controlled exposure.}
Full guidance or interventions enter through defined activation mechanisms; descriptors can already influence selection.
\emph{(ii) Lower context cost.}
Descriptor-first exposure delays full instruction loading until needed.
\emph{(iii) Improved attribution.}
Available, selectable, and active skills can be logged separately.
\emph{(iv) Added coordination burden.}
Descriptors, activation policies, and state transitions must be maintained consistently.

\noindent\textbf{Evidence.}
Progressive disclosure from descriptor exposure to body loading is realised in \href{https://code.claude.com/docs/en/skills#control-who-invokes-a-skill}{Claude Code}, \href{https://docs.openhands.dev/sdk/guides/skill#progressive-disclosure-agentskills-standard}{OpenHands}, \href{https://hermes-agent.nousresearch.com/docs/user-guide/features/skills#progressive-disclosure}{Hermes}, and \href{https://learn.chatgpt.com/docs/build-skills}{OpenAI Codex}. In the PDF example in \Cref{subsec:externalisation}, metadata makes the skill discoverable before Claude reads \texttt{SKILL.md} and then \texttt{forms.md}. This sequence exposes instructions and references progressively, preserving discoverability while deferring their context cost.

Hosts also differ in how activation begins. Explicit activation includes \href{https://learn.chatgpt.com/docs/build-skills#how-chatgpt-and-codex-use-skills}{Codex's \texttt{/skills} menu or \texttt{\$skill-name} invocation}. Policy-mediated activation appears in \href{https://docs.openhands.dev/overview/skills/keyword}{OpenHands' keyword-triggered skills}, which load when a prompt contains a configured trigger. In \href{https://code.claude.com/docs/en/skills#control-who-invokes-a-skill}{Claude Code}, descriptions guide agent-proposed selection; configured skill preloading can instead activate named skills at subagent startup~\citep{anthropic2026claudesubagents}.

\subsection{Skill--Execution Authority Separation}
\label{sec:pattern_skillauthority}

\begin{figure}
    \centering
    \includegraphics[width=\linewidth]{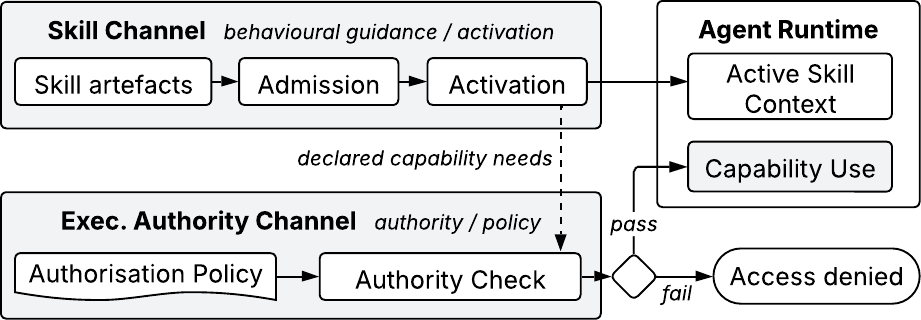}
    \caption{Skill--Execution Authority Separation: host policy determines action permissions and can interpret skill declarations as permission inputs. The check applies to each requested capability use; declared needs are one input alongside policy.}
    \label{fig:skill_authority_separation}
\end{figure}

\noindent\textbf{Intent.}
Separate skill guidance and capability declarations from the host's determination of action permissions (\Cref{fig:skill_authority_separation}).

\noindent\textbf{Context and problem.}
Skills often refer to tools, APIs, MCP servers, scripts, services, or files as task resources, and hosts can interpret skill declarations as permission inputs. Execution authority may also exist independently of any active skill. The architectural problem is to relate a skill's capability requirements and declarations to the permissions applicable to each action.

\noindent\textbf{Forces.}
\emph{(i) Capability expression vs.\ authority granting.}
A skill artefact may need to express expected capabilities, but treating prose references as permission grants confuses stated needs with authorised use.
\emph{(ii) Portable guidance vs.\ contextual authority.}
Skill artefacts should be reusable across agents and workspaces, but tools, credentials, approval rules, and side-effect policies are context-specific.
\emph{(iii) Independent change vs.\ runtime alignment.}
Skill artefacts and authorisation rules may change through different channels, creating runtime mismatches between what a skill expects and what the agent is authorised to do.

\noindent\textbf{Solution.}
Separate skill-exposed behavioural guidance from the policy decision governing capability use. A skill artefact may describe requirements or carry pre-approval declarations, which the host interprets alongside agent, workspace, platform, user, and resource rules. At action time, the host determines the applicable permissions, including the scope and lifetime of any skill-derived grant. It permits, blocks, or escalates the requested capability use accordingly.

\begin{sloppypar}
\noindent\textbf{Variants.}
Variants differ in where the skill--authority separation is resolved.
In \emph{skill-scoped pre-approval}, host-recognised declarations attached to a skill contribute permission grants with a host-defined scope and lifetime. In \emph{agent- or profile-scoped binding}, reusable skills operate under an agent, user, workspace, or profile policy. In \emph{platform-mediated binding}, capability access is governed through platform or administrator controls. A research-informed variant is \emph{verification\hyp{}conditioned binding}, which applies stricter treatment according to skill verification state or side-effect class. Skill-carried restrictions form another variant: Claude Code's \texttt{disallowed-tools} removes tools for the invoking turn, while invoked skill hooks can persist through the session~\citep{anthropic2026claudeskills}.
\end{sloppypar}

\noindent\textbf{Consequences.}
\emph{(i) Portable guidance.}
Skills can describe capability needs while effective permissions are governed in the deployment context.
\emph{(ii) Independent authority governance.}
Tool and action permissions can be granted, narrowed, revoked, and audited independently.
\emph{(iii) Explicit mismatch handling.}
Unavailable or unauthorised capabilities are exposed for runtime handling.
\emph{(iv) Dual-surface maintenance burden.}
The skill artefact's capability metadata and the authority policy require coordinated maintenance.

\begin{sloppypar}
\noindent\textbf{Evidence.}
Skill-scoped pre-approval appears in \href{https://code.claude.com/docs/en/skills#pre-approve-tools-for-a-skill}{Claude Code} and \href{https://docs.github.com/en/copilot/how-tos/copilot-cli/customize-copilot/add-skills}{GitHub Copilot CLI} through \texttt{allowed-tools}. Copilot CLI's documented image-conversion skill illustrates the scope of such a declaration: it packages \texttt{convert-svg-to-png.sh} and can pre-approve \texttt{shell}. This reduces confirmation prompts, but broad shell approval extends the trust decision beyond one conversion script. Accepting the declaration therefore calls for reviewing both the guidance and referenced executable material~\citep{githubCopilotCliSkills}.
\end{sloppypar}

Claude Code's subagents combine preloaded skills with separately configured tool permissions~\citep{anthropic2026claudesubagents}. By contrast, \href{https://opencode.ai/docs/skills/#override-per-agent}{OpenCode's skill-access rules} and \href{https://learn.chatgpt.com/docs/build-skills#optional-metadata}{Codex's invocation policy} govern whether a skill can be loaded or invoked. Research on verifiable artefacts extends authority decisions to verification state and side-effect class, supporting the verification-conditioned variant~\citep{metereSkillsVerifiableArtifacts2026}.

\subsection{Verifiable Skill Contract}
\label{sec:pattern_verifiableskillcontract}

\begin{figure}
    \centering
    \includegraphics[width=\linewidth]{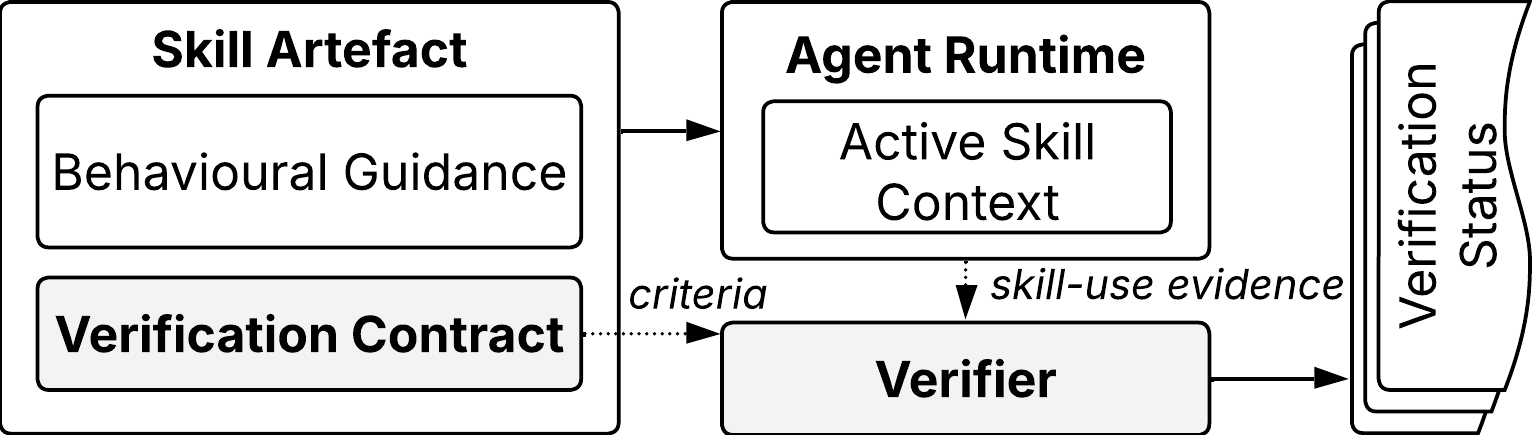}
    \caption{Verifiable Skill Contract: skill-scoped criteria are checked against skill-use evidence by an independent verifier.}
    \label{fig:verifiable_skill_contract}
\end{figure}

\noindent\textbf{Intent.}
Associate a skill artefact with independently checkable criteria for assessing its use and producing an explicit verification status (\Cref{fig:verifiable_skill_contract}).

\noindent\textbf{Context and problem.}
An agent may follow, partially follow, reinterpret, or overextend skill guidance; an executable intervention may run without achieving its intended effect. The architectural problem is to check declared skill-scoped criteria against evidence of the relevant actions, states, or outputs.

\noindent\textbf{Forces.}
\emph{(i) Context flexibility vs.\ verifiability.}
Skills must remain useful across tasks and contexts, but verification requires criteria precise enough to assess enactment.
\emph{(ii) Independent checking vs.\ verifier burden.}
Checking outside the agent's output path reduces self-confirming success claims, but makes verifiers, tests, replay environments, or assertion evaluators part of the trusted architecture.
\emph{(iii) Assurance strength vs.\ contract brittleness.}
Weak criteria provide limited assurance; overly rigid contracts may reject valid context-sensitive behaviour or become costly to maintain.

\noindent\textbf{Solution.}
Associate a verifiable contract with the skill artefact
and introduce a verifier as a separate checking role. The contract specifies the criteria to assess, such as interface constraints, expected state changes, assertions, test cases, or replay checks. The verifier evaluates the corresponding evidence outside the skill's self-report path and produces a status such as \emph{passed}, \emph{failed}, \emph{inconclusive}, or \emph{verifier error}. The status identifies the assessed criteria and is available to downstream decisions.

\noindent\textbf{Variants.}
Variants differ in the checking point and evidence used.
In \emph{boundary checking}, verification occurs at an invocation or interface boundary through typed arguments, schemas, preconditions, postconditions, or output constraints.
In \emph{trajectory checking}, assertions are attached to intermediate steps or states.
In \emph{execution- or replay-based checking}, the skill artefact or candidate skill is run against reference tasks, replay bundles, or regression tests. These offline checks assess the candidate under the tested conditions; checking a current run requires that run's own observations.

\noindent\textbf{Consequences.}
\emph{(i) First-class skill-use status.}
Verification produces a structured status alongside the output or trace.
\emph{(ii) Conditioned composition.}
Downstream processes can condition continuation on verification status.
\emph{(iii) Targeted repair signals.}
Failed clauses or replay diagnostics can guide repair or evolution.
\emph{(iv) Verifier governance burden.}
Contracts and verifiers become artefacts that must themselves be maintained, versioned, and trusted.

\begin{sloppypar}
\noindent\textbf{Evidence.}
Typed arguments for atomic computer-use skills~\citep{chen2026cua} and typed intermediate representations~\citep{ouyang2026skcc} specify structured invocation and representation interfaces. Checking intermediate execution states requires corresponding observations. ContractSkill provides this connection: each web-skill step can carry post-assertions evaluated against page evidence, alongside declared goals, preconditions, postconditions, recovery, and termination checks~\citep{lu2026contractskill}. A failed assertion identifies the step-level criterion that was not met, giving a more specific repair signal than final task failure alone.
\end{sloppypar}

Offline checks before candidate acceptance appear in execution-verified programmatic skill induction~\citep{wang2025inducing} and FlowEvo's interface, safety, and replay validation, with replay checks applied where execution is feasible~\citep{ren2026flowevo}. In the HASP skill-program framework, syntax, interface, mock-execution, and return-type checks similarly assess programmatic candidates before they enter the library~\citep{liu2026harnessing}.

\subsection{Runtime Skill Bill of Materials (Skill-BOM)}
\label{sec:pattern_skillbom}

\begin{figure}
    \centering
    \includegraphics[width=\linewidth]{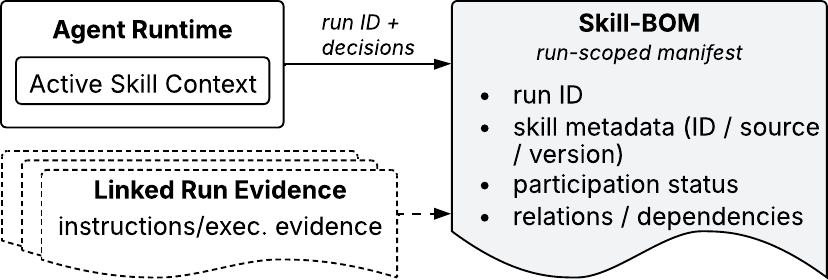}
    \caption{Runtime Skill-BOM: the proposed per-run, skill-centred provenance record links skill artefacts, participation decisions, and run evidence.}
    \label{fig:skillbom}
\end{figure}

\noindent\textbf{Intent.}
For a specific agent run, record the relevant skill artefacts, their source and version metadata, participation decisions, and operational skill-to-skill relations (\Cref{fig:skillbom}).

\noindent\textbf{Context and problem.}
Software Bills of Materials make component provenance and dependency information explicit for supply-chain assurance~\citep{xia2023empirical}.
A skill repository exposes available artefacts, while a run involves particular versions and participation decisions: what was exposed, selected, activated, excluded, unavailable, or used in a composition. Relevant records may be scattered across inventories, configuration, loader events, and execution traces. The architectural problem is to join these records into a run-specific view of identifiable skill artefacts and their participation.

\noindent\textbf{Forces.}
\emph{(i) Run specificity vs.\ inventory overbreadth.}
A repository inventory describes skill supply. Attribution requires identifying the artefacts and decisions associated with the analysed run.
\emph{(ii) Attribution fidelity vs.\ evidence overhead.}
High-fidelity attribution depends on linking evidence across supply, activation, policy, and runtime, but such linkage increases collection and correlation cost.
\emph{(iii) Skill-centred provenance vs.\ audit-log sprawl.}
A complete run log may contain relevant events, but skill attribution remains brittle if skill identities, versions, activation decisions, and relations are scattered across generic traces.

\noindent\textbf{Solution.}
Maintain a run-scoped \emph{Skill-BOM} as an explicit manifest or reconstructed skill-centred record over run evidence. Entries link a run identifier to skill artefact identities, source and version information, and participation events such as exposure, selection, activation, or exclusion. Relevant permission decisions link to the actions and times they govern. Where skill relations affect the run, record the relevant dependency, composition, or load-triggering edges. SkillNet provides related dependency and composition vocabulary~\citep{liang2026skillnet}.

\noindent\textbf{Variants.}
Two implementation options differ in how the record is produced.
In an \emph{explicit run manifest}, the record is stored as a bounded artefact or snapshot.
In a \emph{reconstructed manifest}, the record is assembled from linked loader, activation, configuration, runtime, test, review, or update records through shared run and skill identifiers.

\noindent\textbf{Consequences.}
\emph{(i) Skill-level attribution.}
Developers can distinguish installed skills from skills exposed, activated, excluded, or otherwise relevant to a run.
\emph{(ii) Targeted review and maintenance.}
Findings can be linked to specific skill artefacts, versions, sources, relations, or activation decisions.
\emph{(iii) Provenance burden.}
The pattern requires stable run, skill, version, relation, and decision identifiers. Incomplete records can create false assurance.

\noindent\textbf{Evidence.}
The complete Skill-BOM is a proposed architectural synthesis of documented invocation, identity, integrity, and provenance mechanisms. Artefact identity is supported by skill manifests with stable identifiers for registry indexing~\citep{liang2026skill}, lockfiles and an installation-level agent skill bill of materials~\citep{bhardwaj2026formal}, and provenance records carrying documentation hashes and skill identifiers~\citep{peng2026oxo}. Invocation-time checking connects such records to the loaded content: \citet{shen2026sealing} verify loaded content against audited registry records and, at load time, limit the injected execution envelope to approved manifest permissions within the user's authorised scope.

Invocation records provide the corresponding runtime information. \href{https://docs.snowflake.com/en/user-guide/snowflake-cortex/cortex-agents-skills#monitoring}{Snowflake Cortex Agents} monitors the selected skill, input, and result, while its configuration identifies Snowflake stage or Git sources. Joining an invocation to the resolved skill version brings these records into a run-specific provenance view. Skill-linked trajectory annotations add file-level detail: \citet{liu2026skillsvote} record subtasks with file paths and line ranges. The record can also incorporate \href{https://github.com/aeonfun/aeon/blob/46721b4a2cb64dd91388b39d60d2f6bf40f69991/docs/CORE.md}{Aeon's} per-skill quality scores, failure flags, and health history, and replay or contract evidence bundles~\citep{huang2025audited}.

\subsection{Skill--Agent Co-Evolution Loop}
\label{sec:pattern_skillagentco}

\noindent\textbf{Intent.}
Couple agent execution with governed revision of reusable skill artefacts: accepted skills shape future runs, and run evidence informs changes.

\noindent\textbf{Context and problem.}
Skill artefacts shape runs through activation, while run evidence can inform artefact revision. Missing steps, repeated workarounds, tool failures, user corrections, and successful procedures provide candidates for improvement; \citet{zhao2026thinking} distil successful strategies and failure-mode fixes into reusable skills. A feedback path captures these candidates for refinement and validation: local fixes can overfit a single task or propagate regressions across agents. The architectural problem is to turn run experience into reusable improvements while controlling changes to shared artefacts.

\noindent\textbf{Forces.}
\emph{(i) Cumulative learning vs.\ behavioural stability.}
Skills should improve from accumulated run experience, but shared behavioural artefacts must remain stable enough to be trusted across contexts.
\emph{(ii) Local evidence vs.\ generalisable change.}
Run evidence may reveal a useful procedure, but skill-mediated behaviour depends on specific context and evidence from one context may not generalise.
\emph{(iii) Automation scale vs.\ acceptance cost.}
Frequent skill updates increase manual maintenance effort, while automated proposals still require validation before changing shared guidance.

\begin{figure}
    \centering
    \includegraphics[width=\linewidth]{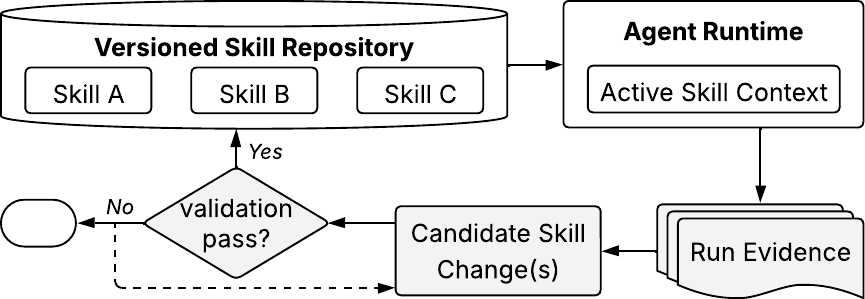}
    \caption{Skill--Agent Co-Evolution Loop: accepted skills shape runs; run evidence produces candidate changes whose acceptance is governed by validation; rejected candidates do not enter the repository.}
    \label{fig:skill_evolution_loop}
\end{figure}

\noindent\textbf{Solution.}
Introduce a validated feedback loop between agent runs and the skill repository (\Cref{fig:skill_evolution_loop}). The forward path activates accepted skill artefacts in later runs. The reverse path turns run evidence into candidate skill creation or refinement. Validate candidates before updating shared artefacts, then version accepted changes and return them through managed supply for future activation.

\begin{sloppypar}
\noindent\textbf{Variants.}
Architectural variants differ mainly in where run-derived skill changes are made reusable.
In \emph{session- or workspace-scoped evolution}, a session, project, or workspace turns its own traces, conversations, failures, or successful procedures into persistent skill changes for later local reuse.
In \emph{repository-governed evolution}, candidate changes are routed through a managed repository for evaluation, deduplication, promotion, or retirement.
In \emph{collective evolution}, evidence from multiple agents, users, or sessions is aggregated before validated updates are synchronised into a shared skill set.
\end{sloppypar}

\noindent\textbf{Consequences.}
\emph{(i) Cumulative improvement.}
Repeated procedures and corrections can become reusable skill artefacts.
\emph{(ii) Reviewable change.}
Candidate changes can be inspected, accepted, or rejected; accepted versions can be rolled back.
\emph{(iii) Validation bottleneck.}
Weak validation promotes drift while strict validation slows useful updates.
\emph{(iv) Evolution provenance burden.}
The loop requires records linking evidence, candidates, validation verdicts, accepted versions, and later runs.

\noindent\textbf{Evidence.}
Feedback-driven skill revision is recurrent, while validation, promotion, and shared governance vary across realisations. Skill1's trajectory distillation and utility updates~\citep{shi2026skill1} illustrate how run experience feeds reusable skills, with related autonomous skill revision from execution experience in XSkill~\citep{jiang2026xskill} and SkillWeaver~\citep{zheng2025skillweaver}.

\begin{sloppypar}
How those changes are accepted varies. FlowEvo compiles successful workflows into validated executable skills~\citep{ren2026flowevo}, whereas Trace2Skill's default path consolidates trajectory-derived patches into a skill directory without validation-based patch selection~\citep{ni2026trace2skill}. Repository governance makes the responsibility for change explicit: SkillOS separates a frozen executor from a curator that edits the skill repository~\citep{ouyang2026skillos}, while \href{https://github.com/aeonfun/aeon/blob/46721b4a2cb64dd91388b39d60d2f6bf40f69991/skills/skill-repair/SKILL.md}{Aeon} routes verification-triggered changes through reviewable pull requests with attached verification plans. Skill-bank operations and retention replay provide further mechanisms for managing reusable changes~\citep{huang2025audited,wuCoEvolvingLLMDecision2026}. The feedback loop can also aggregate experience beyond a single run or user, as in SkillClaw's multi-session evidence and shared skill updates~\citep{ma2026skillclaw}.
\end{sloppypar}

\subsection{Supporting Patterns}
\label{sec:supporting-patterns}

Each entry states the problem, architectural decision, central force, and supporting evidence.

\textbf{Skill Eligibility Gate.}
This pattern places an admission boundary before selectability, upstream of \hyperref[sec:pattern_progressiveskillactivation]{Progressive Skill Activation}: supplied artefacts enter the selectable pool when declared conditions such as validation, trust, enablement, or applicability hold. Because descriptors can steer selection before instructions load, admission governs their exposure. The central force is open reuse versus controlled exposure: a broader pool increases discoverability, while admission criteria restrict which units can participate.
These admission conditions include enable/disable toggles in \href{https://docs.cline.bot/customization/skills#toggling-skills}{Cline} and \href{https://learn.chatgpt.com/docs/build-skills#enable-or-disable-local-codex-skills}{OpenAI Codex}, project trust in \href{https://hermes-agent.nousresearch.com/docs/user-guide/features/skills#trusting-a-project}{Hermes}, and load-time filters in \href{https://docs.openclaw.ai/tools/skills}{OpenClaw}. Where admission depends on validation, checks such as \href{https://docs.praison.ai/docs/cli/skills#check-skills}{PraisonAI's structural and health checks} provide inputs to the decision.

\textbf{Skill Scope Cascade.}
Skill artefacts may exist at multiple scopes, from system or organisation level to project, workspace, user, file-path, or agent level. This pattern resolves which scoped artefacts are visible and, where the host applies precedence, which instance is selected. It specialises configuration resolution for reusable guidance or intervention logic: selecting an unintended artefact can affect subsequent execution. The central force is contextual locality versus governing authority: local skills may better fit the task, while higher-level scopes may encode shared policy, compatibility, or safety constraints.
Source precedence resolves competing artefacts. It appears in enterprise/personal/project precedence in \href{https://code.claude.com/docs/en/skills#resolve-skills-that-share-a-name}{Claude Code}, which also supports namespaced plugin skills and nested discovery; multi-location precedence in \href{https://docs.openclaw.ai/tools/skills#loading-order}{OpenClaw}; global/project precedence in \href{https://docs.cline.bot/customization/skills#where-skills-live}{Cline}; and workspace/global precedence in \href{https://kiro.dev/docs/skills/#skill-scope}{Kiro}. Visibility determines which artefacts a recipient can access, as illustrated by \href{https://cursor.com/docs/skills#skill-directories}{Cursor Agent's project/user directories and nested or path-scoped skills} and \href{https://docs.crewai.com/en/concepts/skills#crew-level-skills}{CrewAI's crew-level skill binding}. A host can combine precedence and visibility when resolving which skills are visible.

\textbf{Skill Resource Materialisation Boundary.}
Skill artefacts may depend on supporting material such as reference files, templates, examples, assets, scripts, or runtime context. Deciding that a skill participates leaves a further decision about which resources to supply and how to obtain them. This pattern makes resource materialisation explicit: reading a reference, resolving an asset, and generating context with a helper have different costs and executable effects. The central force is contextual fidelity versus controlled expansion: task-specific resources can improve the supplied context, while access and execution require appropriate permissions and provenance.
Reference access appears in Claude Code's \href{https://code.claude.com/docs/en/skills#add-supporting-files}{supporting files} and \href{https://docs.cline.bot/customization/skills#bundling-supporting-files}{Cline's referenced resources}, while \href{https://docs.github.com/en/copilot/how-tos/copilot-cli/customize-copilot/add-skills}{Copilot CLI bundles executable scripts}. Materialisation can also occur within invocation: Claude Code's \href{https://code.claude.com/docs/en/skills#inject-dynamic-context}{dynamic context injection} runs commands before sending the rendered skill content to the model. Here, supplying context involves executable operations, connecting \emph{Progressive Skill Activation} to the capability controls of \emph{Skill--Execution Authority Separation}.

\textbf{Skill Extension Bundle.}
Skills can depend on executables, integration code, hooks, MCP-server configuration, or other settings governed through separate review and authority channels. This pattern brings guidance and supporting material into a deployable bundle, so review covers both instructions that direct execution and the code they invoke. It applies across runs, complementing the per-run \textit{Skill Resource Materialisation Boundary}. Separately supplied dependencies require compatible versions and explicit references. The central force is deployment cohesion versus review-surface expansion: bundles simplify sharing and coordinated updates, while admission and trust review cover a broader unit.
Co-delivery is realised through plugins that package skills with hooks, MCP server configuration, and other components in \href{https://code.claude.com/docs/en/plugins#understand-what-a-plugin-is}{Claude Code} and \href{https://developers.openai.com/plugins/build/plugins#plugin-structure}{OpenAI Codex}, and through plugin or extension skills in \href{https://docs.openclaw.ai/tools/skills#plugins-and-skills}{OpenClaw}, \href{https://qwenlm.github.io/qwen-code-docs/en/users/features/skills/#extension-skills}{Qwen Code}, and \href{https://goose-docs.ai/docs/guides/context-engineering/using-skills/#skills-from-plugins}{Goose}.

\textbf{Runtime Skill Repair.}
A skill-in-use may fail verification or skill-scoped enactment criteria during a run. The architectural question is whether execution can continue through a bounded change to its current bindings, composition, or runtime representation. This pattern permits same-run repair constrained by declared checks, locality, and a repair budget. The adapted object is skill-in-use, while the stored artefact and its version remain unchanged. The central force is run continuity versus behavioural preservation: repair changes the current execution arrangement while preserving unaffected verified work. It can be triggered by \textit{Verifiable Skill Contract}.
GraSP provides direct evidence for bounded execution-graph repair: patches preserve graph validity and unaffected verified nodes, and newly introduced nodes use library skills~\citep{xia2026grasp}. SkillTracer provides related evidence for selective repair of attributed plan graphs and also supports persistent refinement~\citep{li2026skilltracer}. Its \href{https://liyuuuuy.github.io/SkillTracer/}{documented address-editing case} expects a \texttt{combobox} for state selection but encounters a \texttt{textbox}; the failed node is localised, the valid prefix preserved, and execution resumed after repair and re-verification.

Beyond these patterns, the evidence base contains broader harness mechanisms that support skill-mediated runs, such as action approval, sandboxing, workflow composition, plugin and package installation, and lifecycle hooks. The patterns specialise them for skill artefacts, and the RA includes those the patterns presuppose as enabling blocks (\Cref{tab:ra-responsibilities}).

\section{Reference Architecture for Skill Harnessing}
\label{sec:reference-architecture}

The RA organises the patterns in \Cref{sec:patterns} and their underlying practices into a system-level Responsibility View (\Cref{fig:refArch}).

\subsection{Overview and Architectural Rationale}
\label{sec:ra-overview}

The system of interest is the skill-harnessing part of an agent system. The agent runtime and execution environment are interfacing systems: the runtime uses skill instructions, resources, or programmatic interventions when producing actions, and the environment performs permitted actions and returns observations.

We define a \emph{Responsibility Viewpoint} for agent-system designers and implementers, using the view and viewpoint concepts of ISO/IEC/IEEE 42010~\citep{isoiecieee42010}. It asks which functions address the concerns in \Cref{subsec:concerns}, how their decisions and outputs depend on one another, and where implementations may vary. Its elements are responsibility blocks grouped into layers, interfacing systems, and cross-cutting responsibilities called substrates. A block represents one responsibility: a related set of decisions or an enabling function; pattern insets mark each pattern's principal allocation. Labelled links express operational interactions or evidence exchange. Concrete components can combine responsibilities or delegate them to shared harness services.

The corresponding \emph{Responsibility View} has four layers, each governing one object: Supply Chain (persistent artefacts, their distribution, and acceptance of revisions), Mediation (runtime participation), Execution Control (skill declarations and execution authority), and Evidence \& Feedback (verification, run-specific provenance, and feedback). Their boundaries separate artefact identity from participation, participation from action permissions, and evidence capture from judgements based on it. Policy and identifiers span all layers and form the two substrates.

\begin{sloppypar}
The activation responsibility illustrates the derivation. Practices from Hermes and OpenAI Codex describe lightweight skill descriptions preceding full instruction loading. They contribute to the practitioner candidate \emph{Progressive Skill Disclosure Loader}, supported by 21 practices from 16 systems. Consolidation with the selection-routing and invocation-policy candidates and academic evidence yielded \emph{Progressive Skill Activation}. With \emph{Skill Resource Materialisation Boundary}, it is allocated to \emph{Activation \& Context Realisation} in Mediation: both decide what the runtime receives, whereas Representation \& Storage retains artefacts and the Execution Control layer permits actions. Because Claude Code's dynamic context injection runs commands while preparing skill content~\citep{anthropic2026claudeskills}, the RA routes these commands through Runtime Action Control.
\end{sloppypar}

The ten patterns have their principal allocation in seven blocks; two further blocks provide enabling functions that these patterns presuppose. Installation, update, sharing, porting, and source-tracking practices motivate Distribution, together with candidate validation and promotion mechanisms~\citep{huang2025audited}. Shared permission and sandbox mechanisms and skill-registered hooks motivate Runtime Action Control, included because skill-initiated operations pass through it, whether a dedicated or a general host gate realises it. Policy coordination and evidence linking motivate the two substrates (\Cref{tab:ra-responsibilities}).

\subsection{Architectural View}
\label{sec:architectural-view}

\begin{figure*}
    \centering
    \includegraphics[width=0.88\linewidth]{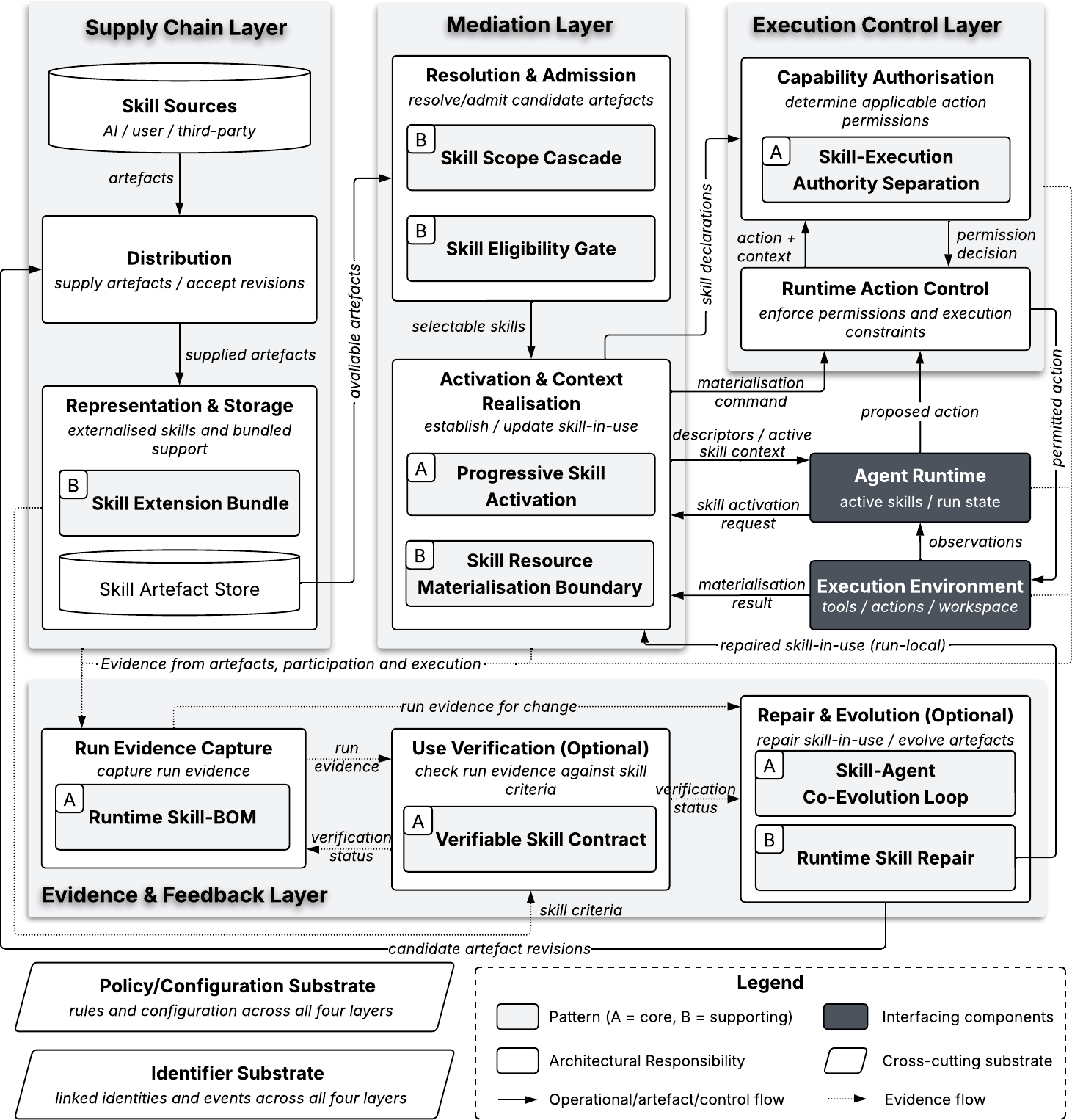}
    \caption{Responsibility View of the reference architecture. Four layers group responsibility blocks by the object they govern; the Policy/Configuration and Identifier substrates apply across all layers. Pattern insets mark each pattern's principal block (A: core; B: supporting). Solid arrows carry operational inputs and outputs; dotted arrows carry evidence, criteria, and results.}
    \label{fig:refArch}
\end{figure*}

\Cref{fig:refArch} presents the Responsibility View. Converging evidence links collect contributions from the supply, participation, and execution responsibilities and from the interfacing systems; the evidence status of each pattern is given in \Cref{tab:catalogue}.

Interactions impose local ordering constraints: an artefact must be accessible before its guidance can participate, and an executable operation must satisfy the applicable controls before it proceeds. Evidence links the decisions and observations produced across these interactions. Components may perform decisions eagerly, lazily, or together, provided the relevant dependencies are preserved. \Cref{tab:ra-responsibilities} summarises the allocation of patterns and enabling functions.

\begin{table*}
\centering
\small
\renewcommand{\arraystretch}{1.12}
\caption{RA responsibility blocks and substrates, derivation basis, and architectural purposes.}
\label{tab:ra-responsibilities}
\begin{tabularx}{\textwidth}{@{}>{\RaggedRight\arraybackslash}p{0.20\linewidth}>{\RaggedRight\arraybackslash}p{0.38\linewidth}>{\RaggedRight\arraybackslash}X@{}}
\toprule
\textbf{Responsibility} & \textbf{Derivation basis} & \textbf{Architectural purpose} \\
\midrule

\rowcolor{tablegray}
Supply Chain:\newline Distribution
& Enabling function. Practices for installing, updating, sharing, porting, and source-tracking skills; acceptance step of Skill--Agent Co-Evolution Loop
& Supplies and updates skill artefacts and governs acceptance of persistent revisions through the required checks or review; retains available source and version information. \\

Supply Chain:\newline Representation \& Storage
& Pattern: Skill Extension Bundle. Practices representing skills as stored packages with metadata and resources
& Retains reusable artefacts and associated resources; bundling is one packaging option. \\

\rowcolor{tablegray}
Mediation:\newline Resolution \& Admission
& Patterns: Skill Scope Cascade; Skill Eligibility Gate
& Resolves scoped candidates and determines which are eligible to participate. \\

Mediation:\newline Activation \& Context Realisation
& Patterns: Progressive Skill Activation; Skill Resource Materialisation Boundary
& Establishes participation through instructions, resources, or programmatic interventions. \\

\rowcolor{tablegray}
Execution Control:\newline Capability Authorisation
& Pattern: Skill--Execution Authority Separation
& Determines applicable action permissions from host and resource policy, including recognised skill declarations and the scope and lifetime of any skill-derived grants. \\

Execution Control:\newline Runtime Action Control
& Enabling function. Practices for permission and sandbox policies and for hooks registered from skill metadata
& Enforces permissions and execution constraints on operations proposed by the runtime or initiated during materialisation. \\

\rowcolor{tablegray}
Evidence \& Feedback:\newline Run Evidence Capture
& Pattern: Runtime Skill-BOM (proposed synthesis)
& Links artefact identities, participation, decisions, actions, and observations within a run. \\

Evidence \& Feedback:\newline Use Verification (Optional)
& Pattern: Verifiable Skill Contract
& Assesses run evidence against artefact-associated or configured skill criteria and records verification status. \\

\rowcolor{tablegray}
Evidence \& Feedback:\newline Repair \& Evolution (Optional)
& Patterns: Runtime Skill Repair; Skill--Agent Co-Evolution Loop
& Routes bounded run-local repair to Mediation and candidate artefact revisions to Distribution for acceptance. \\

Policy/Configuration Substrate
& Synthesis of documented eligibility conditions, invocation policies, permission rules, and repair bounds
& Coordinates rules and configuration consulted across responsibilities. \\

\rowcolor{tablegray}
Identifier Substrate
& Synthesis of documented content hashes, version and provenance records, and invocation identifiers; identifier requirements of Runtime Skill-BOM
& Supplies shared references to artefacts, versions, runs, decisions, actions, observations, and changes. \\

\bottomrule
\end{tabularx}

\par\smallskip
\begin{minipage}{\linewidth}
\footnotesize
\raggedright
Pattern: catalogue pattern allocated to the block. Enabling function: function that pattern solutions presuppose. Optional: selected by requirements (\Cref{sec:useVariationLimits}).
\end{minipage}
\end{table*}

\subsection{Layers and Responsibilities}
\label{sec:ra-layers}

\textbf{Layer 1: Supply Chain.}
\emph{Skill Sources} include AI-generated, user-authored, and third-party artefacts. \emph{Skill Extension Bundle} supplies a packaging option in which guidance and supporting material share a distribution and review boundary. \emph{Distribution} installs and updates artefacts and coordinates the validation and acceptance of persistent revisions, whereas Mediation decides which supplied artefacts become selectable. Where a run must be reproducible or auditable, the identity retained by \emph{Representation \& Storage} must resolve to the artefact revision used.

\textbf{Layer 2: Mediation.}
\emph{Resolution \& Admission} resolves scoped artefacts and determines the selectable pool; skill descriptors can be exposed before activation. \emph{Activation \& Context Realisation} establishes participation through invocation, preloading, or activation conditions, combining \emph{Progressive Skill Activation} with \emph{Skill Resource Materialisation Boundary}. The runtime receives active skill context comprising instructions, resources, or intervention state; programmatic interventions can modify proposed actions. Mediation sends recognised skill declarations to Capability Authorisation, materialisation commands to Runtime Action Control before the content reaches the model, and participation events to Run Evidence Capture.

\textbf{Layer 3: Execution Control.}
\emph{Capability Authorisation} determines action permissions from host and resource policy, including recognised skill declarations. \emph{Runtime Action Control} enforces the decision alongside approval requirements, sandbox constraints, and side-effect controls. It gates operations proposed by the runtime or initiated during resource materialisation. Permitted operations reach the execution environment, whose observations return to the runtime or materialisation process. The two responsibilities can share a host implementation; \emph{Skill--Execution Authority Separation} governs how skill declarations enter the determination.

\textbf{Layer 4: Evidence \& Feedback.}
\emph{Run Evidence Capture} links artefact identities and available version information with participation, permission decisions, actions, and observations. The proposed \emph{Runtime Skill-BOM} integrates these links into an explicit or reconstructed skill-centred record.

Where verification is required, \emph{Use Verification} checks run evidence against skill-scoped criteria through \emph{Verifiable Skill Contract}. Criteria come from Representation \& Storage or the check's configuration. Results return to Run Evidence Capture and can inform Repair \& Evolution; results of checks on candidate revisions can accompany them to Distribution for acceptance.

\emph{Repair \& Evolution} has two change objects and return paths. \emph{Runtime Skill Repair} routes bounded adaptations of current skill-in-use to Mediation, with re-verification and without changing the stored artefact. \emph{Skill--Agent Co-Evolution Loop} sends candidate revisions to Distribution, whose acceptance policy specifies validation before they enter the shared supply.

\subsection{Cross-Cutting Substrates}

The \emph{Policy/Configuration Substrate} supplies the rules consulted for admission, acceptance, activation, authorisation, execution, verification, repair, and evidence retention. These rules can reside in different components while referring to consistent identities and scopes. Reconstructing a past decision additionally requires retaining the applicable policy state or an identifiable policy version with its evidence.

The \emph{Identifier Substrate} joins records about artefacts, versions, runs, participation, actions, checks, and changes. Shared references let evidence capture, verification, and feedback retrieve related records across mechanisms. The required granularity follows the question: identifying an invoked skill differs from reconstructing the revision and permission decision governing an action.

\subsection{Use and Variation}
\label{sec:useVariationLimits}

To apply the RA, an architect identifies the task requirements, allocates documented host mechanisms to responsibilities, and follows their inputs, outputs, and evidence links.

Consider an audit requirement to explain why a commit command was permitted. Claude Code's documented \texttt{commit} skill provides a concrete allocation: its \texttt{SKILL.md} is the stored artefact; invocation and content loading realise Mediation; and the host permission mechanism realises authorisation and action control~\citep{anthropic2026claudeskills}. Following this interaction through the RA identifies the evidence needed: the artefact revision, invocation, applicable permission inputs, and action-time decision. The loading record shows which skill participated, whereas the action-time decision record explains why the command was permitted. For this audit requirement, the design choice is to link action-level permission evidence to participation records, with the additional recording and retention cost this entails.

Variation points concern both the responsibilities required and their realisation. Resolution can apply source precedence, scoped visibility, or both; eligibility can rest on validation, trust review, enablement, or environmental conditions; and resources can be bundled, referenced lazily, or generated by scripts. Explicit invocation gives users direct activation control; descriptor-based selection supports agent-proposed activation; configured preloading supplies named skills or managed instructions loaded in every run. An explicit Skill-BOM makes a run record directly available, while reconstruction joins retained events when needed and depends on sufficient identifiers and event coverage. Use Verification applies when the system must check skill-scoped criteria. Run-local repair applies when bounded adaptation within a run is permitted; persistent evolution applies when reusable artefacts can be revised. A concrete architecture selects these options while preserving the relevant authority, evidence, and change dependencies.

\section{Reference Architecture Evaluation: Mapping and Application}
\label{sec:evaluation}

Following Step~6 (\Cref{sec:methodology-ra}), we evaluate the RA's correspondence with documented mechanisms through cross-system mapping and its application to concrete requirements through focused technical applications. Galster and Avgeriou discuss mapping and prototyping as evaluation approaches and, for RAs grounded in preceding architectures, emphasise instantiation support~\citep{galster2011empirically}.

\subsection{Evaluation Design}
\label{sec:evaluation-design}

The mapping retains eight systems used in RA construction: Claude Code, OpenAI Codex, GitHub Copilot CLI, Hermes Agent, Aeon, OpenClaw, \href{https://learn.microsoft.com/en-us/agent-framework/agents/skills}{Microsoft Agent Framework}, and Snowflake Cortex Agents. We selected them purposively for maximum variation in ecosystem position (coding agents, open-source runtimes, feedback-driven skill systems, enterprise frameworks, and cloud platforms), evidence depth, and responsibility profile, prioritising first-party detail and non-redundant responsibility coverage while avoiding close-lineage variants. Mapping evidence was assessed in October 2026 using first-party documentation and implementation evidence, including documentation beyond the derivation records.

R denotes documented realisation of the architectural purpose stated in \Cref{tab:ra-responsibilities}; P denotes partial support, with part of that purpose or a connection unestablished; N denotes a responsibility not established in the reviewed evidence. Shared harness mechanisms and dedicated skill mechanisms are rated by the same functional rules, with implementation location reported in \Cref{sec:evaluation-results}. A superscript~c marks realisation conditional on a particular channel, optional workflow, or non-default configuration. Distribution concerns deliberate adoption or updating of supplied artefacts. Persistent evolution additionally requires a documented workflow connecting operational or evaluation-run feedback, a candidate revision, review or evaluation for its intended use, and adoption. Evidence for the policy and identifier substrates is assessed within the mapped responsibilities and interactions. The first author rated each system against these rules, recording source locators and a rationale for each cell.

The applications examine execution control in Deep Agents and revision acceptance in Hermes. We selected applications with a documented engineering concern, inspectable mechanisms for the relevant responsibilities, observable success and failure conditions, and a local test path. Deep Agents, LangChain's open-source agent harness, extends the assessment beyond the review corpus. Each protocol stated requirements, RA allocations, and expected observations before its recorded checks. The RA supplied the allocations, state distinctions, and checks; documentation and source inspection identified the host mechanisms; execution established their behaviour. In Deep Agents, a deterministic model client fixed the skill read and proposed action; in Hermes, the test driver submitted candidate revisions through the native \texttt{skill\_manage} tool. Scripted approval or rejection isolated the controls. The checks ran unmodified Deep Agents SDK~0.7.25 and Hermes at pinned commits~\citep{deepagents2026application,hermes2026application} on disposable fixtures, comparing expected observations with native tool results, prompt contents, and resulting file or skill states. Exploratory diagnostics, run before the protocol was fixed, informed the final Deep Agents configuration and expected observations; two exploratory follow-up Hermes checks compared previews and inspected native change records.

\subsection{Cross-System Mapping}
\label{sec:evaluation-results}

Across \Cref{tab:ra-instantiation}, the first six responsibilities are realised in 47 of 48 cells; the 11 R\textsuperscript{c} ratings depend on reviewed, pinned, or interactively confirmed supply channels for Distribution and on configured approval checks or an enabled sandbox for authorisation and control. The remaining P concerns Distribution in Microsoft Agent Framework, rated for the plain framework agent rather than its Harness Agent variant: its source loading provides no adoption or update channel. Representation \& Storage and Activation \& Context Realisation used skill mechanisms in all eight systems, and Resolution \& Admission used skill mechanisms alone in seven and combined with shared access controls in Snowflake. Runtime Action Control used shared harness mechanisms throughout; Capability Authorisation and Run Evidence Capture used shared harness mechanisms alone or combined with skill mechanisms.

\begin{table*}[t]
\centering
\scriptsize
\setlength{\tabcolsep}{3pt}
\renewcommand{\arraystretch}{1.12}
\caption{Documented responsibility realisations in the selected systems.}
\label{tab:ra-instantiation}
\begin{tabularx}{\textwidth}{@{}p{0.22\textwidth}*{9}{>{\centering\arraybackslash}X}@{}}
\toprule
& \multicolumn{2}{c}{\textbf{Supply Chain}} & \multicolumn{2}{c}{\textbf{Mediation}} & \multicolumn{2}{c}{\textbf{Execution Control}} & \multicolumn{3}{c}{\textbf{Evidence \& Feedback}} \\
\cmidrule(lr){2-3}\cmidrule(lr){4-5}\cmidrule(lr){6-7}\cmidrule(lr){8-10}
\textbf{System} & \textbf{Dist.} & \textbf{Rep.} & \textbf{Res./Ad.} & \textbf{Act./Ctx.} & \textbf{Auth.} & \textbf{Ctrl.} & \textbf{Evid.} & \textbf{Verif.} & \textbf{Persist. evol.} \\
\midrule
Claude Code               & R\textsuperscript{c} & R & R & R & R & R & R\textsuperscript{c} & R\textsuperscript{c} & R\textsuperscript{c} \\
OpenAI Codex              & R\textsuperscript{c} & R & R & R & R & R & R\textsuperscript{c} & R\textsuperscript{c} & R\textsuperscript{c} \\
GitHub Copilot CLI        & R\textsuperscript{c} & R & R & R & R & R & R & N & R\textsuperscript{c} \\
Hermes Agent              & R\textsuperscript{c} & R & R & R & R & R & P & N & R\textsuperscript{c} \\
Aeon                      & R & R & R & R & R & R\textsuperscript{c} & P & P & R\textsuperscript{c} \\
OpenClaw                  & R\textsuperscript{c} & R & R & R & R\textsuperscript{c} & R\textsuperscript{c} & P & N & P \\
Microsoft Agent Framework & P & R & R & R & R\textsuperscript{c} & R\textsuperscript{c} & P & P & N \\
Snowflake Cortex Agents   & R\textsuperscript{c} & R & R & R & R & R & P & P & N \\
\bottomrule
\end{tabularx}

\par\smallskip
\begin{minipage}{\linewidth}
\footnotesize
\raggedright
Dist.: Distribution; Rep.: Representation \& Storage; Res./Ad.: Resolution \& Admission; Act./Ctx.: Activation \& Context Realisation; Auth.: Capability Authorisation; Ctrl.: Runtime Action Control; Evid.: Run Evidence Capture; Verif.: Use Verification; Persist.\ evol.: persistent-evolution branch of Repair \& Evolution (run-local repair was not established for any system).\\
R: realised; R\textsuperscript{c}: realised under a stated condition; P: partial; N: not established in the reviewed evidence.
\end{minipage}
\end{table*}

Within Capability Authorisation, Claude Code, Copilot CLI, and Aeon recognise skill declarations; Hermes registers skill-declared environment variables for controlled passthrough to sandboxes, and OpenClaw can allowlist skill-referenced executables on execution nodes. Codex, Microsoft Agent Framework, and Snowflake establish host or resource controls without a documented skill-declaration input.

The clearest variation concerns evidence, verification, and acceptance of change. Claude Code and Codex can link skill participation, permission decisions, and tool results through shared prompt or turn identifiers in configured telemetry; Copilot CLI retains related session events. These joins realise Run Evidence Capture. Revision-level reconstruction, as a Skill-BOM requires, needs more: the cited Claude Code and Codex event schemas identify skills without their artefact revision, whereas Copilot retains loaded skill content. The remaining five Run Evidence Capture ratings are P because the reviewed records leave part of the participation--decision--observation connection unestablished.

The Use Verification ratings separate skill-specific criteria from general evaluation facilities. Claude Code's optional skill\slash plugin evaluation workflow applies skill-targeted graders to dedicated evaluation runs~\citep{anthropic2026claudeskills}. OpenAI's documented, optional Codex evaluation workflow runs a skill-specific prompt set, checks the retained run trace with scripts, and saves a structured rubric assessment~\citep{openai2026skillEvals}. Aeon, Microsoft Agent Framework, and Snowflake provide relevant evaluation facilities, but the reviewed paths do not establish the complete skill-specific criterion-to-observation link.

Persistent evolution varies in how candidate revisions enter reusable supply. Claude Code's evaluation loop qualifies when a candidate revised using evaluation-run feedback enters supply through a controlled release boundary. Codex qualifies when its \href{https://github.com/openai/skills/blob/main/skills/.system/skill-creator/SKILL.md}{skill-creator} iteration and evaluation workflow feeds a reviewed plugin release~\citep{openai2026skillEvals}. Copilot's \href{https://docs.github.com/en/copilot/how-tos/copilot-cli/use-copilot-cli/chronicle}{\texttt{/chronicle improve}} qualifies through proposed changes to managed repository instructions and user selection before writing (\Cref{subsec:adjacent-forms}). With skill-write approval enabled, Hermes qualifies through its documented review path for targeted replacements; the applications below also examine full rewrites. Aeon qualifies through enabled feedback-led pull requests reviewed before merge. OpenClaw's \href{https://docs.openclaw.ai/tools/skill-workshop}{Skill Workshop} supports feedback-driven changes with write-time validation, version history, and undo; without review before reuse, it gives partial support.

\subsection{Focused Applications}
\label{sec:evaluation-applications}

\Cref{tab:ra-applications} groups the eight Deep Agents and six Hermes checks by RA decision.

\begin{table*}[t]
\centering
\small
\setlength{\tabcolsep}{4pt}
\renewcommand{\arraystretch}{1.15}
\caption{RA decisions, tested conditions, criteria, and observations in the focused applications.}
\label{tab:ra-applications}
\begin{tabularx}{\textwidth}{@{}>{\raggedright\arraybackslash}p{0.2\textwidth}>{\raggedright\arraybackslash}p{0.25\textwidth}>{\raggedright\arraybackslash}p{0.17\textwidth}>{\raggedright\arraybackslash}X@{}}
\toprule
\textbf{RA decision} & \textbf{Tested condition} & \textbf{Criterion} & \textbf{Observation} \\
\midrule
\multicolumn{4}{@{}l}{\textit{Deep Agents: protect a file while keeping the report skill usable}} \\
\addlinespace
Runtime Action Control: file-tool denial and shell approval
& File-tool edit vs.\ ungated shell helper, both targeting the protected file
& Protected file unchanged on each route
& Met for the file tool; not met for the ungated shell, which modified the file \\
\addlinespace
& Gated shell: reject the modifying command; approve the report command
& File unchanged; report produced
& Met: rejection blocked the change; approval produced the report with the file unchanged \\
\addlinespace
Capability Authorisation: distinguish declarations from host permissions
& \texttt{allowed-tools:\allowbreak{} execute} vs.\ \texttt{read\_file} vs.\ no declaration, ungated and gated with rejection
& Same enforcement outcomes with and without declarations
& Met: both declarations reached the prompt without changing enforcement \\
\midrule
\multicolumn{4}{@{}l}{\textit{Hermes: accept persistent revisions}} \\
\addlinespace
Distribution: separate candidate from current artefact
& Stage a valid candidate, then reject it
& Current skill unchanged
& Met: current skill unchanged while pending and after rejection \\
\addlinespace
Distribution: separate approval, validation, and adoption
& Approve a valid vs.\ a malformed candidate
& Valid revision applied; malformed revision blocked by validation
& Met: valid revision applied and readable; malformed content failed validation, current version preserved \\
\addlinespace
Distribution: preview before acceptance
& Preview valid and malformed full-rewrite candidates\textsuperscript{a}
& Preview shows each candidate
& Not met: identical failed previews, yet valid content was applied on approval \\
\addlinespace
Identifier Substrate: link candidate, decision, and revision
& Inspect native records after acceptance\textsuperscript{a}
& Records link all three
& Not met: native ledger recorded revision hashes without candidate or decision links \\
\bottomrule
\end{tabularx}

\par\smallskip
\begin{minipage}{\linewidth}
\footnotesize
\raggedright
Gated: each \texttt{execute} (shell) call requires approval through \texttt{interrupt\_on}. \textsuperscript{a}Includes exploratory follow-up checks (\Cref{sec:evaluation-design}).
\end{minipage}
\end{table*}

\textbf{Deep Agents: participation and execution control.}
A report skill's bundled helper sums the input values 1, 2, and 3; the same helper can also overwrite a separate protected file, which must remain unchanged. The configuration combines a shell-capable default backend with a \texttt{/protected/} filesystem route and a write-deny rule for \texttt{/protected/**}. Deep Agents \href{https://docs.langchain.com/oss/python/deepagents/permissions}{documents} that this rule applies to its built-in file tools; \texttt{execute} reaches the same file through the shell backend. Because the RA routes executable operations through Runtime Action Control, we checked both routes and set \texttt{interrupt\_on} for \texttt{execute}, requiring an approval decision for each call to that tool. With the skill loaded, rejection prevented the modifying command; approval of the report command produced \texttt{total=6} and preserved the protected file.

\begin{sloppypar}
The declaration comparison examines how skill metadata relates to host permissions. With identical skill body, helper, inputs, and proposed commands, both \texttt{allowed-tools:\allowbreak{} execute} and \texttt{allowed-tools:\allowbreak{} read\_file} appeared in the prompt. Without the approval gate, the shell modified the protected file under either declaration; with it, \texttt{allowed-tools:\allowbreak{} execute} did not suppress the approval request. This matches the skill middleware, which renders \texttt{allowed-tools} in the prompt without applying it to permissions.
\end{sloppypar}

\textbf{Hermes: approval and revision acceptance.}
With \texttt{skills.write\_approval=true}, proposed changes entered a pending store separate from the current artefact, and approved changes had to pass structural validation before replacing it. Approval of a valid revision made its exact content readable; approval of malformed content failed structural validation and preserved the current version. This exercises the RA path from candidate revision through Distribution's acceptance to supply: an approval decision and successful acceptance are distinct states.

Checking the preview shown to the approver exposed a preview--apply mismatch for the advertised full-rewrite operation. The preview failed in the same way for valid and malformed candidates, yet approving a valid candidate applied its content: the preview parses the full-content \texttt{patch} as a search-and-replace edit with an empty search string, whereas approval applies it as a full rewrite. Hermes's preview code states that the preview should match what approval applies. The native mutation ledger retained before/after revision hashes but no link to the candidate or approval decision; the test driver assembled that association.

\subsection{Findings and Discussion}
\label{sec:evaluation-findings}
\label{sec:discussion}

\textbf{Evidence and feedback need explicit allocation.} Evidence \& Feedback holds 14 of the 15 P and N ratings in \Cref{tab:ra-instantiation}, 11 of them in Run Evidence Capture and Use Verification, whose patterns have mainly research-informed support; run-local repair was not established in any mapping. Only one of the layer's 24 ratings is an unconditional R, against 36 of the 48 in the other layers, and 20 of all 57 realised ratings depend on a channel, workflow, or configuration. Requirements for run-specific provenance, skill-scoped verification, or governed revision therefore need explicitly allocated and configured mechanisms.

\textbf{A skill's permission effects depend on host semantics and configuration.} The same \texttt{allowed-tools} field pre-approves tools in Claude Code and GitHub Copilot CLI (\Cref{sec:pattern_skillauthority}) but left enforcement unchanged in the tested Deep Agents configuration. Pre-approval alone does not restrict the tool set~\citep{anthropic2026claudeskills}, and \href{https://docs.openclaw.ai/tools/exec-approvals}{OpenClaw's gateway and node defaults} let skill helpers run without approval prompts. Adopting a skill therefore requires checking how the target host interprets its declarations and which controls the deployed configuration enables.

\textbf{Review should follow interactions across responsibilities.} Both applications allocated their requirements to existing RA responsibilities, and both separated documented controls from their effective coverage or operation: the Deep Agents file rule left the shell route open, and the Hermes full-rewrite preview did not match what approval applied. Architecture review should therefore follow each route to a protected resource and each candidate through preview, approval, validation, and later supply, with the Identifier Substrate linking the candidate, decision, and resulting revision.

\section{Threats to Validity}
\label{sec:threats}

We discuss construct, internal, and external validity and reliability~\citep{runeson2009guidelines}.

\subsection{Construct Validity}

Translating vendor-specific descriptions into the skill definitions and mapping rules involves judgement, particularly where shared harness mechanisms implement skill-related functions. Public sources also differ in how much they document about controls, execution records, and feedback, and 50 of the 51 academic papers were assessed in their arXiv versions, 43 of them from 2026. Both affect the visibility of mechanisms and the strength of support for individual patterns. Mapping ratings describe documented realisation; the focused applications test operation.

\subsection{Internal Validity}

The authors developed and applied the RA, and the eight mapped systems belong to the review corpus, which can favour interpretations consistent with the proposed structure. The mapping rules registered 15 partial or unestablished realisations, and Deep Agents tests the RA outside the corpus. Fixed action proposals, candidate revisions, and approval decisions attribute each application outcome to host enforcement or acceptance, independently of model behaviour.

\subsection{External Validity}

Rating frequencies describe the eight purposively selected systems. Searches used skill terminology, and academic searches were restricted to titles from 2025 onward; earlier work and equivalent mechanisms described as instructions or rules can therefore be underrepresented.

Host policies, interfaces, and documentation evolve, so mechanism-level findings attach to the documented or tested versions and configurations. The patterns abstract design decisions from vendor-specific mechanisms, and the RA compares implementations at the level of responsibilities; new kinds of mechanism may require revised pattern solutions or responsibility boundaries.

\subsection{Reliability}

After the jointly assessed pilot, the first author performed the remaining source assessment, synthesis, and mapping; records linking source practices to patterns, responsibilities, and ratings make these interpretive decisions inspectable.

Web search results change over time; the recorded queries, follow-up targets, and source records define the reviewed corpus. Replays of the technical applications with fixed inputs and pinned implementations reproduced the recorded outcomes, including those of the exploratory checks.

\section{Related Work}
\label{sec:related-work}

\subsection{Agent Harness and Harness Engineering}
\label{sec:related-harness}

Agent harness research examines how context, tools, execution, and observation support model-driven action. SWE-agent studies the agent--computer interface, and OpenHands provides sandboxed execution and a shared event stream~\citep{yang2024swe,wang2025openhands}. Tang et al. survey harness engineering across workflows, memory, skill libraries, and multi-agent orchestration, relating harness design to model capabilities and system performance~\citep{tang2026harnesssurvey}. Natural-Language Agent Harnesses expresses harness control logic in editable documents that a shared runtime interprets into calls, gates, and contracts~\citep{pan2026natural}, and Agentic Harness Engineering evolves the harness around a fixed model from observations of components, trajectories, and decisions~\citep{lin2026agentic}. Both make the harness itself editable and evolvable; our patterns govern the participation, effects, and revision of the skill artefacts that a harness supplies.

Architectural work identifies responsibilities that our RA builds on. Lu et al.'s reference architecture for foundation-model agents includes guardrails, a black-box recorder, and an AI bill of materials for provenance~\citep{lu2023responsiblearchitecture}. Liu et al.'s agent pattern catalogue documents forces, trade-offs, and known uses; its registry, adapter, and evaluator patterns cover selection, tool integration, and assessment, with Voyager's skill library among the registry examples~\citep{liu2024agentdesignpatterns}. Our catalogue specialises these concerns for identifiable skill artefacts and their staged participation, and the proposed Runtime Skill-BOM combines the provenance and recording roles for skills, linking artefact revisions, participation events, applicable permission decisions, and execution evidence in a per-run record. AgentArcEval complements such architectures with scenario-based evaluation~\citep{lu2026agentarceval}, applicable to concrete architectures instantiated from the RA.

\subsection{Agent Skills and Skill Harnessing}
\label{sec:related-skills}

Skill research contributes mechanisms, surveys, and empirical evidence. Systems build reusable program and workflow libraries~\citep{wang2024voyager,wang2025awm} and verify and select evolving skills~\citep{zhang2026coevoskills,alzubi2026evoskill}; surveys organise skill representation, acquisition, retrieval, and evolution~\citep{zhou2026comprehensive}. SkillsBench and realistic-usage evaluations examine skill utility across tasks and deployment conditions, counterfactual trace auditing examines how skills affect trajectories~\citep{li2026skillsbench,liu2026well,zhou2026counterfactual}, and security studies document malicious skills, skill-file injection, and credential leakage~\citep{liu2026malicious,schmotz2026skill,chen2026credential}. The RA assigns the controls and evidence these findings call for to admission, permission determination, runtime enforcement, evidence capture, and use verification.

Several accounts develop the architectural relationships between skills and host controls. Zhou et al.'s externalisation review connects skills to memory, protocols, and the harness through selection, permission-checked binding, approval gates, audit records, and lifecycle feedback~\citep{zhou2026externalization}. Xu and Yan describe staged loading and execution-context changes, including tool pre-approval, and propose verification gates and trust tiers for lifecycle governance~\citep{xu2026agentskills}. 
SkillGuard treats skills as permission-bearing artefacts and combines manifests, runtime permission control, and policy enforcement over their context influence and side effects~\citep{pan2026skillguard}.
These accounts establish the coupling between skill participation and host controls. Our RA allocates it to two responsibilities: Capability Authorisation interprets recognised skill declarations within host policy, and Runtime Action Control enforces the resulting decisions on executable operations (\Cref{sec:evaluation-applications}).

Badhe et al. propose a reference architecture organised around nine skill-lifecycle stages, from discovery to security governance, covering permission gating, verification, repair, and both temporary adaptation and persistent evolution~\citep{badhe2026agenticskills}. Our RA organises decisions by the objects they govern and specifies the operational and evidence dependencies between responsibilities: commands initiated during resource materialisation pass through Runtime Action Control, and persistent revisions return to Distribution for acceptance.

\begin{sloppypar}
Jiang et al. catalogue seven skill design patterns, each characterised by representation, representative systems, strengths, and risks~\citep{jiang2026sok}. Their progressive-disclosure and self-evolving-library patterns overlap Progressive Skill Activation and Skill--Agent Co-Evolution Loop, marketplace distribution overlaps Skill Extension Bundle and the RA's Distribution responsibility, and their proposed trust tiers parallel Skill--Execution Authority Separation. Our patterns are derived from documented practices, state forces and consequences, and each has a principal allocation in the RA.
\end{sloppypar}

\section{Conclusion}
\label{sec:conclusion}

This paper presents ten architectural patterns for agent skill harnessing, derived from a multivocal literature review, and a reference architecture that connects decisions about a skill across supply, runtime participation, execution control, evidence, and change. Mappings of eight systems show supply, mediation, and execution control broadly realised, with 14 of the 15 partial or unestablished ratings in evidence, verification, and persistent evolution; where these responsibilities are required, they need explicitly allocated mechanisms. Applying the architecture in Deep Agents and Hermes turned requirements into checks: in Deep Agents, an approval gate closed a shell route that file-tool rules left open; in Hermes, approval and validated acceptance were separate, but the full-rewrite preview did not match the applied revision. The architecture therefore directs review to how hosts interpret skill declarations, which controls cover each executable route, and which records link a revision to its candidate and approval.

\section*{Acknowledgement}
Funding: This work was supported by the Responsible AI Research (RAIR) Centre, Adelaide, Australia.

\section*{Declaration of Generative AI Use}

During the preparation and revision of this manuscript, the authors used \textit{ChatGPT (OpenAI)} and \textit{Claude (Anthropic)} to assist with writing tasks such as language polishing, tightening of phrasing, and consistency checking. After using these tools, the authors carefully reviewed, revised, and edited the text, and take full responsibility for all content of the published article.

\printcredits

\bibliographystyle{model1-num-names}

\bibliography{references}



\end{document}